\documentclass{article}
\PassOptionsToPackage{numbers,compress}{natbib}
\usepackage[sglblindworkshop,final]{neurips_2026}
\usepackage[utf8]{inputenc}
\usepackage[T1]{fontenc}
\usepackage{amsmath,amssymb,amsfonts,amsthm}
\usepackage{graphicx}
\usepackage{booktabs}
\usepackage{multirow}
\usepackage{tabularx}
\usepackage{array}
\usepackage{siunitx}
\usepackage{algorithm}
\usepackage[noend]{algpseudocode}
\usepackage{placeins}
\usepackage{dblfloatfix}
\usepackage{microtype}
\usepackage{xcolor}
\usepackage{hyperref}
\usepackage{url}
\newtheorem{theorem}{Theorem}
\newtheorem{lemma}{Lemma}
\title{Robustness Within a Fairness Budget:\\Parity Frontiers and Equal-Opportunity Reversal}
\author{Pablo Rivas \qquad Jeremy Frank\\
Department of Computing Technology, Marist University\\
Poughkeepsie, NY 12601, USA\\
\texttt{pablo.rivas@marist.edu} \qquad \texttt{jeremy.frank1@marist.edu}}
\workshoptitle{LatinX in AI Research Workshop at NeurIPS 2026}
\hypersetup{pdfauthor={Pablo Rivas and Jeremy Frank},pdftitle={Robustness Within a Fairness Budget: Parity Frontiers and Equal-Opportunity Reversal},pdfsubject={LatinX in AI Research Workshop at NeurIPS 2026},hidelinks}
\begin{document}
\maketitle
\begin{abstract}
Adversarial training protects predictive performance under attack but does not determine how decisions are distributed across protected groups. We formulate binary classification as projected gradient descent (PGD) adversarial-risk minimization under a squared soft demographic-parity (DP) budget, optimized with projected multiplier and one-sided penalty updates. On Adult Income, using the binary \texttt{sex} field supplied by the benchmark for group auditing but excluding it from prediction, we test logistic regression and a one-hidden-layer multilayer perceptron (MLP) across three paired seeds, four PGD strengths, and five budgets. At $\varepsilon=0.3$, PGD training improves balanced accuracy over empirical-risk minimization (ERM) by $7.77$ and $11.45$ percentage points and removes $85.7\%$ and $89.5\%$ of attack-induced growth in hard demographic-parity difference (DPD). At the representative interior value $\tau=0.015$, constrained training cuts DPD relative to PGD training by $52.2\%$ and $49.2\%$, while remaining $3.18$ and $8.34$ balanced-accuracy points above ERM. Across all eight seed-averaged model--attack paths, the sampled budget path strictly orders DPD and accuracy and reverses mean signed equal-opportunity difference (EOpp). A rate theorem gives the exact true-positive-rate shift required for reversal, and the measured group rates satisfy it. Within this controlled feature-space study, robustness and parity admit useful joint operating points, but one fairness metric cannot replace a full group-rate audit.
\end{abstract}
\section{Introduction}
\label{sec:introduction}

Adversarial training can make a classifier harder to fool without changing the group disparity already present on clean data. Conversely, stronger control of one group criterion can redirect another. In our Adult study, on-the-fly PGD training suppresses most attack-induced growth in demographic-parity difference (DPD) but leaves the clean gap nearly unchanged; an explicit parity budget then cuts DPD by about half at an interior operating point while preserving a robust balanced-accuracy margin over empirical-risk minimization (ERM). Tighter budgets, however, reverse which protected group has the higher true-positive rate. We ask: \emph{how can robustness and protected-group parity be optimized in one program when success on one fairness target can redirect another?}

Robustness and fairness act at different levels. Adversarial training minimizes predictive loss against an input-level maximizer~\citep{madry2018}; demographic parity compares positive-prediction rates, whereas equal opportunity compares true-positive rates~\citep{hardt2016}. Tabular threat models must also distinguish mutable numeric fields from categorical, immutable, or relational coordinates~\citep{kireev2023tabular,simonetto2024caa,simonetto2024tabularbench}. A robust objective can therefore stabilize an existing allocation without changing its level, while a parity constraint can alter the conditional rates that define another fairness criterion.

Constrained fair classification is well established~\citep{zafar2017fairness,agarwal2018reductions,cotter2019constraints}. Earlier evidence also showed that adversarial training can itself worsen several fairness measures, while subsequent work examined protected-group disparities under adversarial perturbations~\citep{sooksatra2021adversarialfairness,nanda2021robustness,chai2025alignment}. Chai et al.~\citep{chai2025alignment}, the closest comparison, train on classification-adversarial examples using a relaxed fairness condition and a fixed fairness coefficient. We instead organize the experiment around an explicit squared soft-DP inequality: as its right-hand-side budget tightens, what robust performance remains, how does the smooth condition transfer to thresholded DPD, and how do the underlying group rates move? This distinction matters because satisfying a differentiable fairness surrogate need not guarantee its hard-label counterpart~\citep{lohaus2020relaxed}.

\begin{figure*}[t!]
    \centering
    \includegraphics[width=\textwidth]{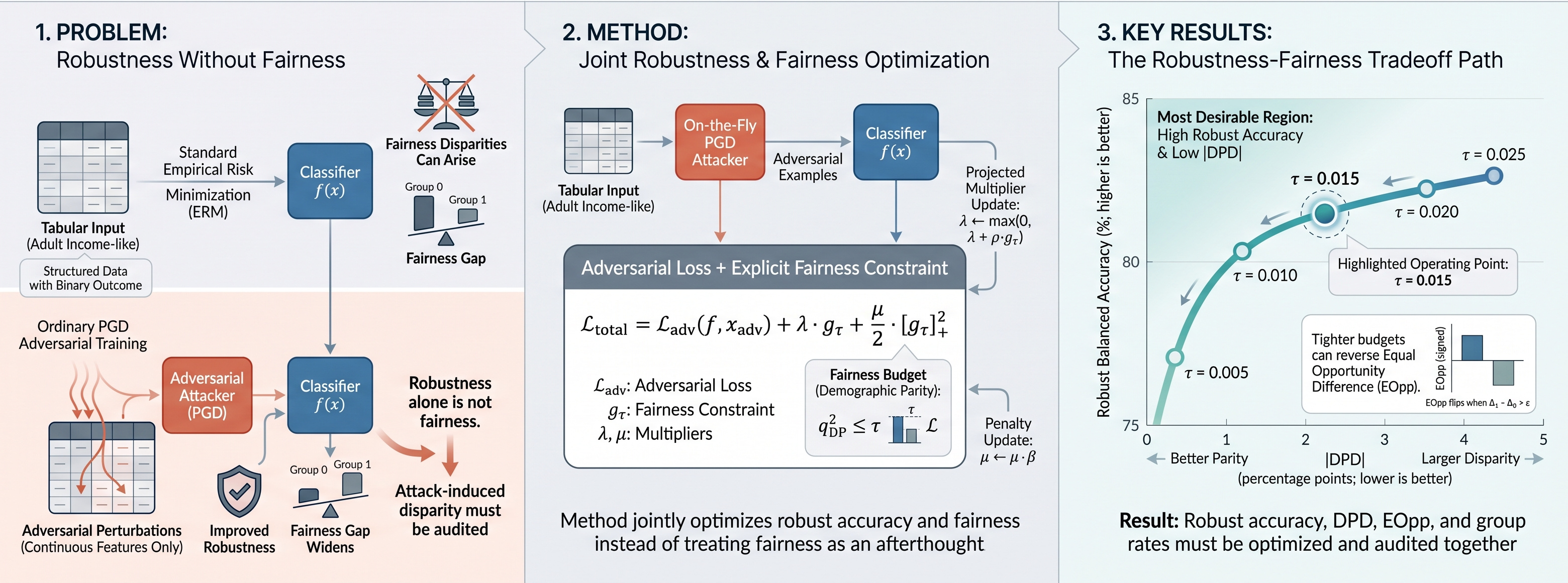}
    \caption{\textbf{Schematic overview.} A parity budget coordinates robust training with group-rate control. Curves, axis values, and group bars are illustrative, not measured results. Section~\ref{sec:method} gives the exact updates; Section~\ref{sec:results} reports the operating paths and EOpp reversal.}
    \label{fig:overview-infographic}
\end{figure*}

We formulate on-the-fly PGD adversarial training under the squared difference between groups' mean predicted probabilities on attacked training examples. Projected multiplier and one-sided penalty updates respond to constraint violation, making $\tau$ a feasibility budget rather than a fixed coefficient. We evaluate the program on Adult Income~\citep{becker1996adult} with logistic regression and a one-hidden-layer MLP, three paired split-and-training seeds, four evaluation strengths, and five budgets. The binary \texttt{sex} field supplied by Adult defines Group~0 and Group~1 for constrained training and auditing but is excluded from prediction. Attacks modify six standardized continuous coordinates; all one-hot coordinates remain fixed. This is a controlled feature-space mechanism study, not a claim of semantic, certified, or universal tabular robustness. Figure~\ref{fig:overview-infographic} summarizes the paper's logic: the robustness--fairness tension, the fairness-budgeted PGD training procedure, and the sampled robustness--parity operating path together with the signed EOpp reversal.

The results identify a useful operating region. At $\varepsilon=0.3$, unconstrained PGD training raises balanced accuracy over ERM by $7.77$ and $11.45$ points while removing $85.7\%$ and $89.5\%$ of attack-induced DPD growth. At the representative interior value $\tau=0.015$, constrained training reduces DPD relative to PGD training by $52.2\%$ and $49.2\%$, yet remains $3.18$ and $8.34$ balanced-accuracy points above ERM. Across all eight seed-averaged model--attack paths, the five tested budgets order hard DPD and balanced accuracy in opposite directions, and mean signed EOpp crosses zero. We use \emph{frontier} descriptively for this sampled five-point path; we do not claim global Pareto optimality or that $\tau=0.015$ is intrinsically best.

\paragraph{Contributions.}
(1) \textbf{Budgeted robust formulation.} We formulate and evaluate PGD adversarial-risk minimization under a squared soft-DP inequality; the solver is an implementation instrument, not a new general optimization claim. (2) \textbf{Controlled operating-path study.} Across two model forms, paired seeds, four attack strengths, and five budgets, we measure the sampled hard-DPD--accuracy path and identify an interior point where both models retain a robust margin over ERM while cutting DPD by about half. (3) \textbf{Cross-metric diagnostic.} We prove that DPD does not determine EOpp direction, derive the exact true-positive-rate shift required for reversal, and show that the measured group rates satisfy it while exposing a leveling-down component.

Robustness, demographic parity, and equal opportunity are linked through learned scores and group rates, but none can stand in for the others. The value of the budget is not a single preferred setting; it is the auditable path and the group-rate account that it makes visible.
\section{Related Work}
\label{sec:related-work}

\paragraph{Tabular adversarial robustness.}
Tabular attacks must account for feature cost, categorical validity, immutability, and relations among fields~\citep{kireev2023tabular,simonetto2024caa,simonetto2024tabularbench}. Those methods support broad domain-valid assessment. Our controlled threat model instead freezes every one-hot coordinate and perturbs six standardized continuous coordinates. This narrower construction keeps the attacked subspace fixed across training regimes and isolates the fairness-budget question; it is not presented as semantic, causal, or actionability-aware robustness.

\paragraph{Fairness as constrained learning.}
Prior work casts fair classification as covariance control, reductions to cost-sensitive learning, or primal--dual optimization of data-dependent constraints~\citep{zafar2017fairness,agarwal2018reductions,cotter2019constraints,cotter2019nondifferentiable}. FairALM applies an augmented Lagrangian to group fairness, while budget-based fair learning uses related optimization machinery~\citep{lokhande2020fairalm,fontana2026fairlab}. We use this established optimization base for adversarial risk under a protected-group soft-DP inequality. Because smooth relaxations can diverge from thresholded metrics~\citep{lohaus2020relaxed}, the full sweep measures hard DPD instead of treating $\tau$ as a hard-label guarantee.

\paragraph{Robust fairness and the object being equalized.} Image-based fair adversarial training often equalizes clean or robust performance across output classes~\citep{xu2021fairrobust,wei2023cfa,zhang2024faal}; the relevant groups are labels, not protected attributes. Sooksatra et al. showed that adversarial training can widen class-wise accuracy and robustness disparities and can also increase an equalized-odds-based unfairness measure on Adult, directly motivating methods that optimize robustness and fairness together rather than auditing them independently~\citep{sooksatra2021adversarialfairness}. Other protected-group studies audit unequal adversarial susceptibility~\citep{nanda2021robustness}, study demographic fairness under attack~\citep{zeng2023demographic}, or combine recourse parity with decision-boundary margin on structured data~\citep{sharma2021fairn}. Chai et al.~\citep{chai2025alignment} are closest to the present formulation: they study protected-group fairness under adversarial perturbations and implement a fixed-weight fairness term. Our contribution is not the existence of fair adversarial training or augmented-Lagrangian fairness. It is the combined object of an explicit swept inequality budget, its hard-DPD versus robust-accuracy path, and a theorem-backed account of signed EOpp reversal.
\section{Fairness-Budgeted Adversarial Training}
\label{sec:method}

We place a soft DP budget inside adversarial training. Unlike a fixed regularization coefficient, $\tau$ states an admissible group-score gap while the multiplier and penalty adapt to violations.

\paragraph{Robust empirical risk.}
Let $f_\theta(x)$ be a logit, $p_\theta(x)=\sigma(f_\theta(x))$, and $h_\theta(x)=\mathbf 1\{p_\theta(x)\geq 1/2\}$. If $\mathcal C$ indexes perturbable continuous coordinates, define
$\mathcal U_\varepsilon=\{\delta:\delta_j=0\ \forall j\notin\mathcal C,\ \|\delta_{\mathcal C}\|_\infty\leq\varepsilon\}$.
The ideal robust risk and finite-step minibatch loss are
\begin{align*}
\mathcal R_{\mathrm{rob}}(\theta;\varepsilon)
&=\mathbb E\!\left[\max_{\delta\in\mathcal U_\varepsilon}
\ell_{\mathrm{BCE}}(f_\theta(X+\delta),Y)\right],\\
\widehat{\mathcal L}_{\mathrm{adv}}(\theta;\mathcal B)
&=\frac{1}{|\mathcal B|}\sum_{i\in\mathcal B}
\ell_{\mathrm{BCE}}(f_\theta(\widetilde x_i^{\mathrm{tr}}),y_i),
\end{align*}
where ten-step PGD approximates the inner maximum and attacked inputs are detached before the outer update~\citep{madry2018}.

\paragraph{Soft parity budget and updates.}
For an attacked sample set $\mathcal E=\{(w_i,y_i,a_i)\}_{i=1}^n$ containing both groups, let
\begin{align}
q_{\mathrm{DP}}(\theta;\mathcal E)
&=\frac{1}{n_0}\sum_{i:a_i=0}p_\theta(w_i)
 -\frac{1}{n_1}\sum_{i:a_i=1}p_\theta(w_i),\notag\\
s_{\mathrm{DP}}(\theta;\mathcal E)&=q_{\mathrm{DP}}(\theta;\mathcal E)^2,
\qquad g_\tau(\theta;\mathcal E)=s_{\mathrm{DP}}(\theta;\mathcal E)-\tau.
\label{eq:soft-dp-budget}
\end{align}
Equation~\eqref{eq:soft-dp-budget} makes the budget explicit: feasibility is $g_\tau\leq0$, equivalently $|q_{\mathrm{DP}}|\leq\sqrt\tau$, so $\tau$ has squared probability-gap units. Since training constrains scores before thresholding, it does not formally bound hard DPD.

Within an epoch, the parameter update minimizes
\begin{equation}
\widehat{\mathcal J}_\tau
=\widehat{\mathcal L}_{\mathrm{adv}}
+\lambda g_\tau+\frac{\mu}{2}[g_\tau]_+^2,
\qquad [u]_+=\max(u,0).
\label{eq:aug-objective}
\end{equation}
The one-sided term in Equation~\eqref{eq:aug-objective} penalizes violations without rewarding over-satisfaction. A minibatch containing only one group uses adversarial BCE alone because its group gap is undefined. At epoch end, the method regenerates the attacked training split, computes $g_k$, and applies
\begin{align}
\lambda_{k+1}&=[\lambda_k+g_k]_+,\notag\\
\mu_{k+1}&=
\begin{cases}
2\mu_k,& g_k>0\ \text{and}\ g_k>\tfrac14[g_{k-1}]_+,\\
\mu_k,&\text{otherwise},
\end{cases}
\label{eq:multiplier-penalty-update}
\end{align}
with $\lambda_0=0$ and $\mu_0=1$. Equation~\eqref{eq:multiplier-penalty-update} therefore raises feasibility pressure only when the measured residual warrants it. The saved model is the validation-feasible epoch with minimum adversarial validation BCE. This is the executed stochastic nonconvex procedure; no global-convergence or component-superiority claim is made.

\paragraph{What DP control implies for equal opportunity.}
For a fixed clean or attacked input $W$, write $\pi_a=\Pr(Y=1\mid A=a)$, $\mathrm{PPR}_a=\Pr(h_\theta(W)=1\mid A=a)$, $\mathrm{TPR}_a=\Pr(h_\theta(W)=1\mid A=a,Y=1)$, and $\mathrm{FPR}_a=\Pr(h_\theta(W)=1\mid A=a,Y=0)$. Define $\mathrm{DPD}=\mathrm{PPR}_0-\mathrm{PPR}_1$ and $\mathrm{EOpp}=\mathrm{TPR}_0-\mathrm{TPR}_1$.

\paragraph{Contribution boundary.}
The projected multiplier and penalty updates are established solver components used for this budgeted program; we do not claim a new general optimizer or superiority over fixed-weight training. Likewise, Theorem~\ref{thm:dp-eopp-direction} is a metric diagnostic rather than a convergence result: it characterizes what hard DPD can and cannot identify and gives an exact condition that can be checked against observed group rates. It does not make the smooth training budget a formal hard-DPD guarantee.

\begin{theorem}[DP does not determine equal-opportunity direction]
\label{thm:dp-eopp-direction}
Assume every group--label cell has positive probability. Then:
\textbf{(i)}
\begin{align}
\mathrm{DPD}
&=\pi_0\mathrm{EOpp}
 +(1-\pi_0)(\mathrm{FPR}_0-\mathrm{FPR}_1)\notag\\
&\quad +(\pi_0-\pi_1)(\mathrm{TPR}_1-\mathrm{FPR}_1).
\label{eq:dpd-rate-decomposition}
\end{align}
\textbf{(ii)} For every $u\in[-1,1]$, a valid binary distribution exists with $\pi_0=\pi_1=1/2$, $\mathrm{DPD}=0$, and $\mathrm{EOpp}=u$; hence $\sup\{|\mathrm{EOpp}|:\mathrm{DPD}=0\}=1$.
\textbf{(iii)} Compare two operating points on the same population. If $e=\mathrm{EOpp}^{(0)}>0$ and $\Delta_a=\mathrm{TPR}_a^{(1)}-\mathrm{TPR}_a^{(0)}$, then
\begin{equation}
\mathrm{EOpp}^{(1)}<0
\quad\Longleftrightarrow\quad
\Delta_1-\Delta_0>e.
\label{eq:eopp-reversal-condition}
\end{equation}
Equality yields zero EOpp.
\end{theorem}

\begin{proof}
Conditioning the positive-prediction event on $Y$ gives
$\mathrm{PPR}_a=\pi_a\mathrm{TPR}_a+(1-\pi_a)\mathrm{FPR}_a$.
Subtracting the two group identities and adding and subtracting
$\pi_0\mathrm{TPR}_1+(1-\pi_0)\mathrm{FPR}_1$ yields~\eqref{eq:dpd-rate-decomposition}.
For (ii), let $t=|u|$ and use equal base rates. Choose
$(\mathrm{TPR}_0,\mathrm{TPR}_1,\mathrm{FPR}_0,\mathrm{FPR}_1)=(t,0,0,t)$ for $u\geq0$ and $(0,t,t,0)$ for $u<0$. Both groups then have $\mathrm{PPR}=t/2$ and EOpp $u$; assigning positive mass to each group and label cell constructs a valid joint law. For (iii), substitution gives $\mathrm{EOpp}^{(1)}=e+\Delta_0-\Delta_1$, which is negative exactly under~\eqref{eq:eopp-reversal-condition}.
\end{proof}

The theorem is diagnostic rather than an optimization guarantee: it turns an observed sign change into a falsifiable group-rate condition and shows why success on DP cannot identify EOpp direction. Appendix~\ref{app:theory-extensions} expands the rate algebra, and Appendix~\ref{app:full-algorithm} gives the complete executed update order.
\section{Experimental Setup}
\label{sec:experimental-setup}

\paragraph{Data and models.}
We use OpenML Adult (data id 1590), derived from the University of California, Irvine (UCI) Adult dataset~\citep{becker1996adult}. Complete-case filtering leaves 45,222 records; $Y=1$ denotes annual income above \$50,000. The supplied binary \texttt{sex} field defines $A=0$ (Male) and $A=1$ (Female), is excluded from prediction, and is used only for constrained training and group evaluation. This benchmark encoding does not represent nonbinary or intersectional identities. For each seed $r\in\{0,1,2\}$, we create a label-stratified 60/20/20 train/validation/test split and fit preprocessing on the training split alone. Six continuous fields are standardized; seven categorical fields are one-hot encoded, giving 102 coordinates (6 continuous and 96 one-hot). The predictors are logistic regression and a one-hidden-layer MLP with 64 rectified linear unit (ReLU) activations.

\paragraph{Training and checkpointing.}
We compare ERM, on-the-fly PGD training~\citep{madry2018}, and fairness-constrained PGD training. All use binary cross-entropy (BCE) with logits, Adam with learning rate $10^{-4}$, batch size 32, at most 40 epochs, patience 10, and threshold $0.5$. ERM selects minimum clean validation BCE; PGD training selects minimum adversarial validation BCE; Fair PGD selects the minimum adversarial validation BCE among validation-feasible epochs. We sweep $\tau\in\{0.005,0.010,0.015,0.020,0.025\}$. The value $\tau=0.015$ is a representative interior point used for exact comparisons, not a test optimum; the complete sweep supports the operating-path claims.

\paragraph{Threat model and evaluation.}
Only \texttt{age}, \texttt{fnlwgt}, \texttt{education-num}, \texttt{capital-gain}, \texttt{capital-loss}, and \texttt{hours-per-week} may change; every one-hot coordinate remains fixed. Thus $\varepsilon$ is measured in standardized feature units. Training uses ten clean-start sign steps with nominal radius $\varepsilon_{\mathrm{tr}}=0.3$ and step size $\alpha_{\mathrm{tr}}=0.01$, without coordinatewise range clipping. Because the path begins at the clean input, its maximum coordinate displacement is $10(0.01)=0.10$ even though projection uses the nominal $0.3$ box. We thus treat it as the fixed model-current training stressor, not evidence that the $0.3$ inner maximum is solved. Final evaluation attacks each trained model with ten-step PGD at $\varepsilon\in\{0,0.1,0.2,0.3\}$, one seeded random start, $\alpha_{\mathrm{ev}}=2\varepsilon/10$, and coordinatewise bounds learned from the transformed training split. The untargeted attack maximizes true-label BCE, not a fairness objective. Final robustness statements concern this one-start continuous feature-space evaluation PGD; they are neither certificates nor claims that the training and evaluation attacks coincide.

\paragraph{Metrics and replication.}
Balanced accuracy is primary and area under the receiver operating characteristic curve (AUC) is secondary. Attack-success rate (ASR) is the fraction of clean-correct predictions that become incorrect after attack. DPD and EOpp follow the signed definitions in Section~\ref{sec:method}; positive values mean Group~0 has the higher named rate. Each seed changes the split, preprocessing fit, initialization, minibatch order, and seeded attack realization, so method comparisons are paired within seed. We report test-set means $\pm$ sample standard deviations over the three complete replicates, without normality, confidence-interval, or significance-test claims. Appendix~\ref{app:data-contract} gives the full field and perturbation contract; Appendix~\ref{app:executed-settings} records every executed setting.
\section{Results}
\label{sec:results}

\begin{figure*}[!b]
  \centering
  \includegraphics[width=\textwidth]{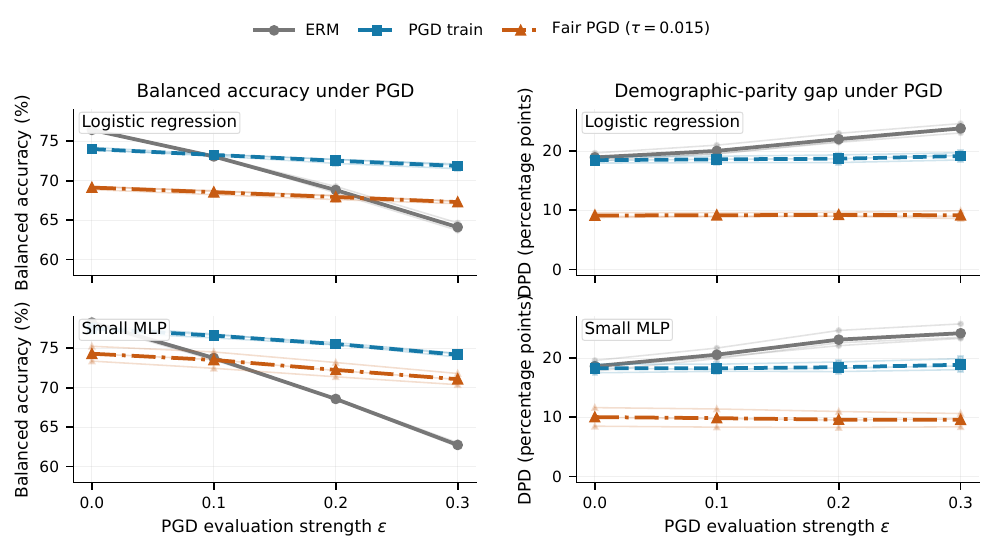}
  \caption{\textbf{Robust training stabilizes performance; the parity budget lowers DPD.} Large markers are three-seed means and faint traces are individual seeds. PGD training limits attack-induced balanced-accuracy loss and DPD growth but begins from nearly the same clean DPD. Fair PGD at $\tau=0.015$ lowers DPD across the evaluated range and remains above ERM in robust balanced accuracy at $\varepsilon=0.3$. Here $\varepsilon=0$ is clean evaluation and $\mathrm{DPD}=\mathrm{PPR}_0-\mathrm{PPR}_1$.}
  \label{fig:robustness-disparity}
\end{figure*}

\paragraph{PGD exposes a coupled performance--parity vulnerability in ERM.}
Across the three paired seeds, increasing the evaluation radius from clean input to $\varepsilon=0.3$ lowers ERM balanced accuracy by 12.34 percentage points for logistic regression and 15.52 points for the MLP. Over the same interval, hard DPD rises by 4.91 and 5.58 points, and ASR reaches 14.33\% and 18.64\%. Every model--seed pair follows both directions. The attack therefore reduces predictive utility while widening the positive-prediction gap already present on clean data. Figure~\ref{fig:robustness-disparity} summarizes this paired degradation and contrasts it with the trajectories produced by robust and constrained training.

\paragraph{On-the-fly PGD training stabilizes both accuracy and disparity under attack.}
The PGD-training trajectories remain comparatively flat as $\varepsilon$ increases. At $\varepsilon=0.3$, PGD training raises balanced accuracy over ERM by 7.77 points for logistic regression and 11.45 points for the MLP, and lowers ASR by 11.96 and 15.35 points. It reduces attack-induced DPD growth from 4.91 to 0.70 points and from 5.58 to 0.58 points---reductions of 85.7\% and 89.5\%. Yet clean DPD changes only from 18.95 to 18.48 points and from 18.64 to 18.30 points. Robust training therefore resists attack-driven disparity amplification while leaving the original parity level as a separate target. Table~\ref{tab:key-results} reports the exact clean and strongest-attack operating points for all three training regimes. Appendix~\ref{app:complete-central-results} gives the full attack-strength record, and Appendix~\ref{app:seed-level-results} resolves the strongest attack by seed.

\begin{table*}[t]
\centering
\caption{\textbf{Exact operating points confirm a joint robust--fair region.} Three-seed mean $\pm$ sample standard deviation. Balanced accuracy (BA) and attack success rate (ASR) are percent; DPD and signed EOpp are percentage points. Fair PGD uses $\tau=0.015$; ASR is omitted for clean input.}
\label{tab:key-results}
\footnotesize
\sisetup{separate-uncertainty=true,table-number-alignment=center}
\begin{tabular*}{\textwidth}{@{\extracolsep{\fill}}lll S[table-format=2.2(3)] S[table-format=2.2(3)] S[table-format=-2.2(3)] S[table-format=2.2(3)]@{}}
\toprule
Evaluation & Model & Training & {BA} & {DPD} & {EOpp} & {ASR} \\
\midrule
Clean & Logistic & ERM & 76.45(12) & 18.95(69) & 12.21(479) & {--} \\
 &  & PGD train & 74.02(19) & 18.48(52) & 17.32(427) & {--} \\
 &  & Fair PGD & 69.14(22) & 9.13(44) & -2.97(437) & {--} \\
 & Small MLP & ERM & 78.27(25) & 18.64(91) & 7.99(463) & {--} \\
 &  & PGD train & 77.53(39) & 18.30(75) & 8.74(486) & {--} \\
 &  & Fair PGD & 74.32(94) & 10.04(157) & -8.70(538) & {--} \\
\midrule
PGD $\varepsilon=0.3$ & Logistic & ERM & 64.12(48) & 23.85(79) & 13.98(478) & 14.33(68) \\
 &  & PGD train & 71.88(31) & 19.18(62) & 16.06(442) & 2.37(26) \\
 &  & Fair PGD & 67.30(18) & 9.17(67) & -2.74(509) & 1.94(3) \\
 & Small MLP & ERM & 62.75(26) & 24.22(137) & 6.42(507) & 18.64(28) \\
 &  & PGD train & 74.20(27) & 18.89(95) & 9.87(499) & 3.30(67) \\
 &  & Fair PGD & 71.09(71) & 9.59(113) & -9.79(454) & 3.03(8) \\
\bottomrule
\end{tabular*}
\end{table*}

\paragraph{The constrained procedure yields feasible checkpoints throughout the budget grid.}
All 30 constrained runs reach validation feasibility, defining a checkpoint for every model. Twenty-seven complete all 40 epochs; three MLP runs stop after obtaining feasible candidates. These histories establish usable checkpoints throughout the grid, not convergence to a local or global optimum.

\begin{figure*}[!b]
  \centering
  \includegraphics[width=\textwidth]{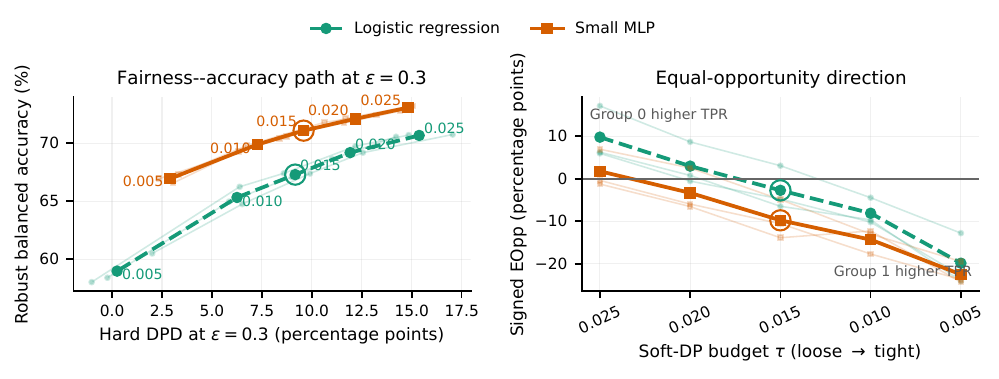}
  \caption{\textbf{The DP budget traces a sampled operating path and can reverse equal-opportunity direction.} Large markers are three-seed means; faint marks and traces are individual seeds. \textbf{Left:} tighter budgets lower hard DPD with a corresponding robust-accuracy cost at $\varepsilon=0.3$; outlined points mark $\tau=0.015$ as an interior illustration, not an optimum. \textbf{Right:} mean signed EOpp changes direction as $\tau$ tightens, while seed variation shows that no single crossing budget applies across models and runs. Lines connect tested budgets only; ``frontier'' denotes this sampled path.}
  \label{fig:budget-frontier-eopp}
\end{figure*}

\paragraph{The full sweep traces a sampled seed-averaged operating path.}
Figure~\ref{fig:budget-frontier-eopp}(a) orders all five budgets at $\varepsilon=0.3$. For logistic regression, tightening $\tau$ from 0.025 to 0.005 reduces mean hard DPD from 15.37 to 0.25 points while balanced accuracy moves from 70.67\% to 58.98\%; for the MLP, DPD moves from 14.81 to 2.91 points while balanced accuracy moves from 73.07\% to 66.95\%. Both metrics share this mean ordering in all eight model--attack paths. At run level, 94/96 adjacent balanced-accuracy comparisons and 92/96 adjacent DPD comparisons follow the mean order. \emph{Frontier} denotes this sampled five-point path, not global Pareto optimality. Appendix~\ref{app:full-budget-paths} reports every model--attack budget grid and its AUC path.

\paragraph{A representative interior budget demonstrates a joint operating region.}
We inspect $\tau=0.015$ as an interior point, not an optimum. At $\varepsilon=0.3$, Fair PGD lowers DPD relative to unconstrained PGD training by 10.01 points for logistic regression and 9.30 points for the MLP, corresponding to reductions of 52.2\% and 49.2\%. The balanced-accuracy costs relative to PGD training are 4.59 and 3.11 points, but the constrained models remain 3.18 and 8.34 points above ERM, with ASR of 1.94\% and 3.03\% (Table~\ref{tab:key-results}). DPD falls in all 24 paired model--seed--evaluation comparisons, and the robust margin over ERM at $\varepsilon=0.3$ holds in all 6 paired seed comparisons. These comparisons establish useful robust--fair operating points within the tested PGD procedure, not superiority over fairness-constrained ERM or fixed-weight alternatives.

\paragraph{Tighter DP control reverses mean EOpp, but not at one universal threshold.}
Figure~\ref{fig:budget-frontier-eopp}(b) shows a mean direction change in all eight model--attack paths between $\tau=0.025$ and $0.005$. At $\varepsilon=0.3$, logistic EOpp moves from $+9.76$ to $-19.87$ points and MLP EOpp from $+1.75$ to $-22.50$ points. Near the crossing, however, between-seed variation exceeds the mean: logistic EOpp is $+2.95\pm4.98$ at $\tau=0.020$ and $-2.74\pm5.09$ at $0.015$, while MLP EOpp is $+1.75\pm4.50$ at $0.025$ and $-3.34\pm5.08$ at $0.020$. These near-zero means do not establish equal opportunity. Crossings occur in 18/24 individual-seed paths, with locations that depend on model and seed. Table~\ref{tab:group-mechanism} reports the clean-input PPR and TPR shifts used to test the theorem.

\begin{table*}[!b]
\centering
\caption{\textbf{Group-rate shifts explain the EOpp reversal.} Clean-input paired changes from PGD training to Fair PGD at $\tau=0.015$, reported as mean $\pm$ sample standard deviation in percentage points. Changes are constrained minus unconstrained; $e$ is EOpp before the constraint.}
\label{tab:group-mechanism}
\footnotesize
\sisetup{separate-uncertainty=true,table-number-alignment=center}
\textit{Paired group-rate changes}\\
\begin{tabular*}{\textwidth}{@{\extracolsep{\fill}}l S[table-format=-2.2(3)] S[table-format=-2.2(3)] S[table-format=-2.2(3)] S[table-format=-2.2(3)]@{}}
\toprule Model & {$\Delta\mathrm{PPR}_0$} & {$\Delta\mathrm{PPR}_1$} & {$\Delta\mathrm{TPR}_0$} & {$\Delta\mathrm{TPR}_1$} \\ \midrule
Logistic & -8.39(20) & 0.95(15) & -15.75(32) & 4.54(134) \\
Small MLP & -6.44(103) & 1.82(23) & -11.20(190) & 6.23(139) \\
\bottomrule\end{tabular*}\par\smallskip
\textit{Exact EOpp reversal check}\\
\begin{tabular*}{\textwidth}{@{\extracolsep{\fill}}l S[table-format=2.2(3)] S[table-format=2.2(3)] S[table-format=-2.2(3)]@{}}
\toprule Model & {$e=\mathrm{EOpp}_{\rm before}$} & {$\Delta_1-\Delta_0$} & {$\mathrm{EOpp}_{\rm after}$} \\ \midrule
Logistic & 17.32(427) & 20.30(109) & -2.97(437) \\
Small MLP & 8.74(486) & 17.43(59) & -8.70(538) \\
\bottomrule\end{tabular*}
\end{table*}

\paragraph{The group-rate mechanism satisfies the theorem and exposes leveling down.}
Table~\ref{tab:group-mechanism} compares unconstrained PGD training with Fair PGD at $\tau=0.015$ on clean inputs. Logistic Group~0 PPR falls by 8.39 points while Group~1 rises by 0.95; for the MLP, the changes are $-6.44$ and $+1.82$ points. Most DPD closure therefore comes from lowering the initially higher-rate group, with a smaller increase in the lower-rate group. At the level of seed means, the TPR changes satisfy Eq.~\eqref{eq:eopp-reversal-condition}: the relative shifts are $20.30\pm1.09>17.32\pm4.27$ points and $17.43\pm0.59>8.74\pm4.86$ points, producing final EOpp values of $-2.97\pm4.37$ and $-8.70\pm5.38$. The same strict inequality holds in all eight seed-averaged model--attack comparisons at this budget. The DP objective succeeds on its stated target, while the asymmetric rate movement shows why aggregate parity must be interpreted with group PPR, TPR, FPR, and utility. Appendix~\ref{app:group-results} supplies the complete group-level record.
\section{Discussion, Limitations, and Impact}
\label{sec:discussion}

\paragraph{The budget makes the operating path auditable.}
Figure~\ref{fig:robustness-disparity} and Table~\ref{tab:key-results} separate attack stability from baseline parity: PGD training protects performance but leaves clean DPD nearly unchanged. The explicit constraint changes that disparity level. Figure~\ref{fig:budget-frontier-eopp}(a) exposes the cost across five sampled, seed-averaged points; ``frontier'' does not assert Pareto optimality. An externally chosen performance floor and disparity ceiling identify acceptable tested points, while $\tau=0.015$ remains an interior illustration.

\paragraph{Metric choice belongs in the design problem.}
Theorem~\ref{thm:dp-eopp-direction} is an exact audit rule, not an optimization theorem. Equation~\eqref{eq:dpd-rate-decomposition} decomposes DPD into TPR, FPR, and base-rate terms; Eq.~\eqref{eq:eopp-reversal-condition} gives the necessary-and-sufficient TPR-shift condition for reversal. Figure~\ref{fig:budget-frontier-eopp}(b) and Table~\ref{tab:group-mechanism} confirm that the measured rates satisfy it. Most DPD reduction comes from lowering the initially higher-rate group's positive decisions, with a smaller increase for the other group. This leveling-down component makes lower disparity an incomplete welfare criterion~\citep{mittelstadt2024leveling}; DPD, EOpp, their component rates, and utility require joint assessment.

\paragraph{Controlled scope and attribution limits.}
Adult's binary \texttt{sex} field supports a controlled mechanism study, not claims about nonbinary or intersectional groups. The linear and nonlinear models and paired seeds test consistency within that setting. Other tabular datasets, protected attributes, and group distributions would test generalization. Three seeds do not locate a precise EOpp crossing, and one-start PGD over six standardized numeric coordinates tests feature-space rather than semantic or certified robustness. Two targeted controls remain important: fairness-constrained ERM (Fair-ERM) and PGD training with a fixed fairness weight $\lambda$. Matched splits and evaluation strengths would show whether adversarial training or adaptive fairness enforcement shifts the DPD--accuracy path or EOpp crossing. The current results characterize the budgeted PGD procedure, not its superiority over those alternatives. Historical wall time and peak memory were not retained, precluding efficiency claims; the theorem, unlike the empirical comparisons, is dataset-independent.

\paragraph{Responsible interpretation.}
Excluding the protected field from prediction prevents direct use at inference, not proxy effects. Adult is a historical benchmark, not a deployment study. Selecting $\tau$ in practice requires an account of benefits, errors, relevant group rates, and credible attacks. The useful robust--parity points found here must therefore be judged by their group-rate profiles. Appendices~\ref{app:attack-comparison} and~\ref{app:static-control} give the attack and training controls; Appendix~\ref{app:artifact-map} records artifact provenance.
\section{Conclusion}
\label{sec:conclusion}

Under a controlled Adult feature-space protocol, this study establishes that an explicit soft-DP budget can be integrated into adversarial training to produce useful joint robustness--parity operating points. On-the-fly PGD training improves robust balanced accuracy and suppresses most attack-induced DPD growth, while revealing that the clean disparity remains a distinct optimization target. The five-budget sweep makes the price of tighter parity explicit rather than burying it inside a fixed coefficient. At the representative interior point $\tau=0.015$, both models remain above ERM in robust balanced accuracy while reducing the DPD of unconstrained PGD training by about half. 

The sweep also exposes a deeper cross-metric effect: tighter budgets redistribute group true-positive rates enough to reverse signed EOpp precisely when $\Delta_1-\Delta_0>e$, as established by the theorem. This reversal is the central diagnostic result. Robust accuracy, DPD, EOpp, and the group rates that compose them are coordinated but non-interchangeable objectives. A credible fair--robust model should therefore be selected from the full operating path and evaluated through both aggregate disparities and group-specific rates, rather than judged by one tolerance or one fairness scalar.

\section*{Acknowledgments}

The authors thank the members of the Rivas.AI Lab (\url{https://lab.rivas.ai}) for their support and valuable feedback throughout the project, and the anonymous reviewers for their useful comments.

\bibliographystyle{plainnat}
\bibliography{references}
\clearpage
\appendix
\section{Expanded rate analysis}
\label{app:theory-extensions}

The main paper studies two signed group differences under a fixed clean or attacked evaluation condition. Let $A\in\{0,1\}$ be the protected-group indicator, $Y\in\{0,1\}$ the label, and $H\in\{0,1\}$ the hard prediction. For group $a$, define the positive-prediction rate (PPR), true-positive rate (TPR), false-positive rate (FPR), and positive-label prevalence as
\begin{align*}
\mathrm{PPR}_a &= \Pr(H=1\mid A=a), &
\mathrm{TPR}_a &= \Pr(H=1\mid A=a,Y=1),\\
\mathrm{FPR}_a &= \Pr(H=1\mid A=a,Y=0), &
\pi_a &= \Pr(Y=1\mid A=a).
\end{align*}
The demographic-parity difference (DPD) and signed equal-opportunity difference (EOpp) are
\[
\mathrm{DPD}=\mathrm{PPR}_0-\mathrm{PPR}_1,
\qquad
\mathrm{EOpp}=\mathrm{TPR}_0-\mathrm{TPR}_1.
\]
Positive values mean that Group~0 has the larger named rate.

\subsection{Why DPD does not identify EOpp}
Conditioning on $Y$ within group $a$ gives
\begin{align}
\mathrm{PPR}_a
&= \Pr(H=1\mid A=a,Y=1)\Pr(Y=1\mid A=a) \notag\\
&\quad + \Pr(H=1\mid A=a,Y=0)\Pr(Y=0\mid A=a) \notag\\
&= \pi_a\mathrm{TPR}_a+(1-\pi_a)\mathrm{FPR}_a.
\label{eq:app-total-probability}
\end{align}
Subtracting the Group~1 identity from the Group~0 identity yields the direct decomposition in Equation~\eqref{eq:app-dpd-expanded}:
\begin{align}
\mathrm{DPD}
={}&\pi_0\mathrm{TPR}_0+(1-\pi_0)\mathrm{FPR}_0 \notag\\
&-\pi_1\mathrm{TPR}_1-(1-\pi_1)\mathrm{FPR}_1.
\label{eq:app-dpd-expanded}
\end{align}
Adding and subtracting $\pi_0\mathrm{TPR}_1$ and $(1-\pi_0)\mathrm{FPR}_1$ gives
\begin{align}
\mathrm{DPD}
={}&\pi_0\mathrm{EOpp}
 +(1-\pi_0)(\mathrm{FPR}_0-\mathrm{FPR}_1) \notag\\
&+(\pi_0-\pi_1)(\mathrm{TPR}_1-\mathrm{FPR}_1).
\label{eq:app-dpd-decomposition}
\end{align}
Equation~\eqref{eq:app-dpd-decomposition} separates three sources of DPD: the signed TPR difference, the signed FPR difference, and the effect of unequal label prevalences. A change in one term can be offset by the others, so lower DPD alone cannot determine the sign of EOpp. The metrics align under additional structure. If $\pi_0=\pi_1=\pi>0$ and $\mathrm{FPR}_0=\mathrm{FPR}_1$, then
\begin{equation}
\mathrm{DPD}=\pi\,\mathrm{EOpp},
\label{eq:app-alignment}
\end{equation}
Equation~\eqref{eq:app-alignment} shows that the two signed differences then share a direction. The demographic-parity budget imposes neither premise.

\subsection{Constructive non-identification at exact parity}
The non-identification claim admits a direct construction. Fix any $u\in[-1,1]$, set $t=|u|$, and choose equal base rates $\pi_0=\pi_1=1/2$. For $u\ge0$, use
\[
(\mathrm{TPR}_0,\mathrm{TPR}_1,\mathrm{FPR}_0,\mathrm{FPR}_1)=(t,0,0,t),
\]
and for $u<0$, use $(0,t,t,0)$. In both cases, Equation~\eqref{eq:app-total-probability} gives $\mathrm{PPR}_0=\mathrm{PPR}_1=t/2$, hence $\mathrm{DPD}=0$, while $\mathrm{EOpp}=u$. At $t=1$, exact demographic parity coexists with EOpp equal to $1$ or $-1$. Therefore,
\begin{equation}
\sup\{ |\mathrm{EOpp}|:\mathrm{DPD}=0\}=1.
\label{eq:app-nonidentification}
\end{equation}
Equation~\eqref{eq:app-nonidentification} states the resulting identification limit; it does not claim that the experimental models attain these extreme rates.

\subsection{Exact sign-reversal condition}
Let the initial signed EOpp be $e=\mathrm{EOpp}^{(0)}>0$, and let
$\Delta_a=\mathrm{TPR}_a^{(1)}-\mathrm{TPR}_a^{(0)}$. Then
\begin{equation}
\mathrm{EOpp}^{(1)}=e+\Delta_0-\Delta_1.
\label{eq:app-eopp-update}
\end{equation}
Equation~\eqref{eq:app-eopp-update} makes the before-to-after accounting explicit. Consequently,
\begin{equation}
\mathrm{EOpp}^{(1)}<0
\quad\Longleftrightarrow\quad
\Delta_1-\Delta_0>e.
\label{eq:app-reversal}
\end{equation}
Equation~\eqref{eq:app-reversal} is necessary and sufficient. Equality reaches zero; a smaller relative Group~1 shift leaves the final EOpp nonnegative. Appendix~\ref{app:group-results} evaluates this identity using the measured group rates.

\subsection{What projected multiplier accumulation implies}
\begin{lemma}[Average signed residual]
\label{lem:app-average-residual}
Let $\lambda_{k+1}=[\lambda_k+\eta_\lambda g_k]_+$ with $\lambda_0\ge0$ and $\eta_\lambda>0$. For every $K\ge1$,
\begin{equation}
\frac{1}{K}\sum_{k=0}^{K-1}g_k
\le
\frac{\lambda_K-\lambda_0}{\eta_\lambda K}.
\label{eq:app-average-residual}
\end{equation}
\end{lemma}
\begin{proof}
Because $[z]_+\ge z$, the projected update satisfies
$\lambda_{k+1}\ge\lambda_k+\eta_\lambda g_k$. Thus
$g_k\le(\lambda_{k+1}-\lambda_k)/\eta_\lambda$. Summing over $k=0,\ldots,K-1$ telescopes and yields Equation~\eqref{eq:app-average-residual}.
\end{proof}
Lemma~\ref{lem:app-average-residual} controls the average \emph{signed} residual when the multiplier remains bounded. It is not a global convergence theorem and does not bound the average positive violation without further assumptions. The full update order is given next.

\section{Complete constrained-training algorithm}
\label{app:full-algorithm}

Algorithm~\ref{alg:fair-pgd-training} separates the three optimization roles that are easy to conflate. Projected gradient descent (PGD) searches for classification-adversarial inputs; the primal update reduces adversarial binary cross-entropy (BCE) under the current feasibility pressure; and epoch-level multiplier and penalty updates react to the full-training residual. The inner attack does not maximize a fairness objective, and gradients do not pass through attack generation.

\begin{algorithm*}[!htbp]
\caption{Fairness-constrained on-the-fly PGD adversarial training}
\label{alg:fair-pgd-training}
\small
\begin{algorithmic}[1]
\Require training and validation splits $D_{\mathrm{tr}},D_{\mathrm{val}}$; budget $\tau$; maximum epochs $K$; patience $P$
\State initialize model parameters $\theta$, multiplier $\lambda_0\gets0$, penalty $\mu_0\gets1$, and previous violation $v_{-1}\gets+\infty$
\State best feasible validation loss $L_\star\gets+\infty$; smallest violation $v_\star\gets+\infty$; stall counter $c\gets0$
\For{epoch $k=0,\ldots,K-1$}
  \For{minibatch $B=\{(x_i,y_i,a_i)\}\subset D_{\mathrm{tr}}$}
    \State generate $x_i^{\mathrm{adv}}\gets\mathsf{PGD}_{\mathrm{tr}}(\theta,x_i,y_i)$ for all $i$ and detach attacked inputs
    \State $L_{\mathrm{adv}}\gets |B|^{-1}\sum_i\ell_{\mathrm{BCE}}(f_\theta(x_i^{\mathrm{adv}}),y_i)$
    \If{both protected groups occur in $B$}
      \State $q_B\gets\operatorname{mean}_{a_i=0}p_\theta(x_i^{\mathrm{adv}})-\operatorname{mean}_{a_i=1}p_\theta(x_i^{\mathrm{adv}})$
      \State $g_B\gets q_B^2-\tau$
      \State $J_B\gets L_{\mathrm{adv}}+\lambda_k g_B+\frac{\mu_k}{2}[g_B]_+^2$
    \Else
      \State $J_B\gets L_{\mathrm{adv}}$ \Comment{the group-score gap is undefined}
    \EndIf
    \State update $\theta$ by one Adam step on $J_B$
  \EndFor
  \State regenerate attacked full training and validation splits with the current model
  \State $g_k\gets s_{\mathrm{DP}}(\theta;D_{\mathrm{tr}}^{\mathrm{adv}})-\tau$ and $v_k\gets[g_k]_+$
  \State compute adversarial validation BCE $L_k^{\mathrm{val}}$ and validation surrogate $s_k^{\mathrm{val}}$
  \If{$s_k^{\mathrm{val}}\le\tau$ and $L_k^{\mathrm{val}}<L_\star$}
    \State save the feasible checkpoint; $L_\star\gets L_k^{\mathrm{val}}$; $c\gets0$
  \ElsIf{$s_k^{\mathrm{val}}>\tau$ and $s_k^{\mathrm{val}}-\tau<v_\star$}
    \State retain the least-violating in-memory fallback; $v_\star\gets s_k^{\mathrm{val}}-\tau$
  \Else
    \State $c\gets c+1$
  \EndIf
  \State $\lambda_{k+1}\gets[\lambda_k+g_k]_+$
  \If{$g_k>0$ and $g_k>\tfrac14 v_{k-1}$}
    \State $\mu_{k+1}\gets2\mu_k$
  \Else
    \State $\mu_{k+1}\gets\mu_k$
  \EndIf
  \State stop when $c\ge P$
\EndFor
\State \Return persisted validation-feasible checkpoint with minimum adversarial validation BCE
\end{algorithmic}
\end{algorithm*}
\FloatBarrier

The reported evaluations use only persisted validation-feasible checkpoints. Every one of the 30 constrained runs attains feasibility at least once, so this selection rule is defined throughout the full model--budget grid. Twenty-seven runs complete all 40 epochs; three MLP runs stop after producing feasible candidates. The algorithm is an executed stochastic nonconvex procedure, not a claim of global optimality or superiority over all fixed-weight solvers.

\section{Dataset, preprocessing, and perturbation contract}
\label{app:data-contract}

\paragraph{Population and splits.}
We use OpenML Adult data id 1590, derived from the University of California, Irvine Adult benchmark~\citep{becker1996adult}. Complete-case filtering retains 45,222 records. The binary target is annual income above \$50,000. For each seed $r\in\{0,1,2\}$, a label-stratified split assigns 60\% to training, 20\% to validation, and 20\% to testing, yielding 27,132, 9,045, and 9,045 records. Each seed changes the split, training-fitted preprocessor, initialization, minibatch order, and seeded attack randomness; it is therefore a complete paired replicate rather than only a model-init seed.

\paragraph{Protected field.}
The supplied binary \texttt{sex} field defines Group~0 as Male and Group~1 as Female. It is retained for constrained training and group-level evaluation but removed from the predictive feature matrix. This prevents direct use of the field at inference; it does not remove proxy structure from the remaining covariates, and the benchmark encoding does not represent nonbinary or intersectional identities.

\paragraph{Predictor and transformation.}
The six continuous fields are \texttt{age}, \texttt{fnlwgt}, \texttt{education-num}, \texttt{capital-gain}, \texttt{capital-loss}, and \texttt{hours-per-week}. The seven categorical fields are \texttt{workclass}, \texttt{education}, \texttt{marital-status}, \texttt{occupation}, \texttt{relationship}, \texttt{race}, and \texttt{native-country}. Continuous fields are standardized and categorical fields are one-hot encoded using transformations fitted on training data only. The resulting model input has 102 coordinates: six continuous followed by 96 one-hot coordinates.

\paragraph{Perturbation set.}
Let $\mathcal C$ index the six continuous transformed coordinates. For input $x$,
\begin{equation}
\mathcal S_\varepsilon(x)=\{z:z_j=x_j\ \forall j\notin\mathcal C,\ \|z_{\mathcal C}-x_{\mathcal C}\|_\infty\le\varepsilon\}.
\label{eq:app-threat-set}
\end{equation}
Training PGD uses this set without clipping to observed training ranges. Evaluation intersects it with coordinatewise minima and maxima computed from the transformed training split. All one-hot coordinates remain fixed, and $\varepsilon$ is measured in standardized units rather than raw feature units.

\paragraph{Scope.}
The mask in Equation~\eqref{eq:app-threat-set} prevents fractional one-hot entries but does not impose integer, causal, actionability, or cross-field consistency constraints. For example, an attacked standardized \texttt{education-num} value may move while the one-hot \texttt{education} category remains fixed. The study therefore measures continuous feature-space robustness, not semantic, actionable, causal, or certified tabular robustness.

\section{Executed settings and checkpoint rules}
\label{app:executed-settings}

Table~\ref{tab:app-hyperparameters} records the complete executed protocol. Its most important distinction is between training and final evaluation. Training PGD uses ten clean-start steps of size $0.01$ inside a nominal radius of $0.3$; the path can therefore move a coordinate by at most $0.10$. Final evaluation instead uses one seeded random start, ten steps of size $2\varepsilon/10$, and training-derived coordinate bounds. Robustness claims concern the specified evaluation attack, not an assertion that the nominal training inner maximum is solved.

\begin{table*}[!htbp]
\centering
\caption{\textbf{Complete executed protocol.} Training and evaluation PGD differ in initialization, step size, and range clipping.}
\label{tab:app-hyperparameters}
\footnotesize
\begin{minipage}[t]{0.49\textwidth}
\textit{Data, models, and fitting}\par
\begin{tabularx}{\linewidth}{@{}>{\raggedright\arraybackslash}p{0.31\linewidth}>{\raggedright\arraybackslash}X@{}}
\toprule
Setting & Executed value \\
\midrule
Dataset & OpenML Adult id 1590; complete cases; 45,222 records \\
Group field & \texttt{sex}; Group 0 = Male, Group 1 = Female; excluded from predictor \\
Splits & Label-stratified 60/20/20 for seeds 0, 1, 2 \\
Preprocessing & Six standardized continuous fields; seven one-hot categorical fields; training-fit only \\
Input dimension & 102: 6 continuous and 96 one-hot coordinates \\
Models & Logistic regression; one-hidden-layer MLP with 64 ReLU units \\
Optimization & BCE with logits; Adam; $10^{-4}$ learning rate; batch size 32 \\
Training horizon & At most 40 epochs; patience 10; decision threshold 0.5 \\
Reporting & Three paired replicates; mean $\pm$ sample standard deviation \\
Compute path & CPU-compatible reference implementation \\
\bottomrule
\end{tabularx}
\end{minipage}\hfill
\begin{minipage}[t]{0.49\textwidth}
\textit{Attack and fairness optimization}\par
\begin{tabularx}{\linewidth}{@{}>{\raggedright\arraybackslash}p{0.31\linewidth}>{\raggedright\arraybackslash}X@{}}
\toprule
Setting & Executed value \\
\midrule
Training PGD & Nominal $\varepsilon=0.3$; 10 clean-start steps; $\alpha=0.01$; no range clipping \\
Maximum path displacement & $10(0.01)=0.10$ standardized units \\
Evaluation PGD & $\varepsilon\in\{0,0.1,0.2,0.3\}$; 10 steps; $\alpha=2\varepsilon/10$; one random start \\
Attack coordinates & Six continuous coordinates; all one-hot coordinates fixed \\
Evaluation bounds & Coordinatewise train-set minima and maxima \\
Inner objective & True-label BCE; no fairness-aware inner attack \\
Soft-DP surrogate & $q_{\rm DP}^2$, $q_{\rm DP}=\mathbb E[p\mid A=0]-\mathbb E[p\mid A=1]$ \\
Budget grid & $\tau\in\{0.005,0.010,0.015,0.020,0.025\}$ \\
Dual update & $\lambda_{k+1}=[\lambda_k+g_k]_+$; $\lambda_0=0$ \\
Penalty update & $\mu_0=1$; double when $g_k>0$ and $g_k>0.25[g_{k-1}]_+$ \\
Fair checkpoint & Lowest adversarial validation BCE among feasible epochs \\
\bottomrule
\end{tabularx}
\end{minipage}
\end{table*}
\FloatBarrier

The linear model is a single affine logit. The nonlinear comparator is a one-hidden-layer multilayer perceptron (MLP) with 64 rectified linear unit activations. All central models use Adam, batch size 32, threshold $0.5$, and at most 40 epochs. Empirical risk minimization (ERM) selects the lowest clean validation BCE, unconstrained PGD training selects the lowest adversarial validation BCE, and Fair PGD selects the lowest adversarial validation BCE among validation-feasible epochs. The budget values are fixed before inspecting the test set; $\tau=0.015$ is a representative interior operating point, not a test-selected optimum.

\section{Complete central PGD evaluation}
\label{app:complete-central-results}

Table~\ref{tab:app-epsilon-sweep} reports ERM, on-the-fly PGD training, and Fair PGD at $\tau=0.015$ across all four evaluation strengths. The ERM rows show the baseline failure: balanced accuracy (BA) falls, attack-success rate (ASR) rises, and DPD widens as $\varepsilon$ increases. PGD training changes that trajectory for both model classes, while Fair PGD lowers DPD throughout the same range and retains a robust BA margin over ERM at the strongest attack.

\begin{table*}[!htbp]
\centering
\caption{\textbf{Complete central PGD evaluation.} Three-seed mean $\pm$ sample standard deviation. BA and ASR are percent; DPD and signed EOpp are percentage points. Fair PGD uses $\tau=0.015$.}
\label{tab:app-epsilon-sweep}
\footnotesize
\textit{Logistic regression}\par
\begin{tabular*}{\textwidth}{@{\extracolsep{\fill}}l c c c c c@{}}
\toprule
Training & $\varepsilon$ & BA & DPD & EOpp & ASR \\
\midrule
ERM & 0.0 & $76.45\pm0.12$ & $18.95\pm0.69$ & $12.21\pm4.79$ & -- \\
ERM & 0.1 & $73.08\pm0.12$ & $20.05\pm0.84$ & $11.80\pm4.66$ & $3.68\pm0.10$ \\
ERM & 0.2 & $68.82\pm0.42$ & $22.04\pm0.85$ & $14.09\pm5.64$ & $8.56\pm0.51$ \\
ERM & 0.3 & $64.12\pm0.48$ & $23.85\pm0.79$ & $13.98\pm4.78$ & $14.33\pm0.68$ \\
PGD train & 0.0 & $74.02\pm0.19$ & $18.48\pm0.52$ & $17.32\pm4.27$ & -- \\
PGD train & 0.1 & $73.28\pm0.11$ & $18.62\pm0.54$ & $16.94\pm4.40$ & $0.77\pm0.16$ \\
PGD train & 0.2 & $72.54\pm0.29$ & $18.72\pm0.65$ & $16.36\pm4.35$ & $1.59\pm0.25$ \\
PGD train & 0.3 & $71.88\pm0.31$ & $19.18\pm0.62$ & $16.06\pm4.42$ & $2.37\pm0.26$ \\
Fair PGD ($\tau{=}0.015$) & 0.0 & $69.14\pm0.22$ & $9.13\pm0.44$ & $-2.97\pm4.37$ & -- \\
Fair PGD ($\tau{=}0.015$) & 0.1 & $68.55\pm0.27$ & $9.17\pm0.38$ & $-3.44\pm4.30$ & $0.58\pm0.08$ \\
Fair PGD ($\tau{=}0.015$) & 0.2 & $67.94\pm0.29$ & $9.23\pm0.39$ & $-3.31\pm4.30$ & $1.24\pm0.11$ \\
Fair PGD ($\tau{=}0.015$) & 0.3 & $67.30\pm0.18$ & $9.17\pm0.67$ & $-2.74\pm5.09$ & $1.94\pm0.03$ \\
\bottomrule
\end{tabular*}
\medskip
\textit{Small MLP}\par
\begin{tabular*}{\textwidth}{@{\extracolsep{\fill}}l c c c c c@{}}
\toprule
Training & $\varepsilon$ & BA & DPD & EOpp & ASR \\
\midrule
ERM & 0.0 & $78.27\pm0.25$ & $18.64\pm0.91$ & $7.99\pm4.63$ & -- \\
ERM & 0.1 & $73.77\pm0.23$ & $20.59\pm0.97$ & $8.08\pm5.23$ & $5.26\pm0.06$ \\
ERM & 0.2 & $68.58\pm0.11$ & $23.14\pm1.36$ & $7.50\pm5.96$ & $11.63\pm0.16$ \\
ERM & 0.3 & $62.75\pm0.26$ & $24.22\pm1.37$ & $6.42\pm5.07$ & $18.64\pm0.28$ \\
PGD train & 0.0 & $77.53\pm0.39$ & $18.30\pm0.75$ & $8.74\pm4.86$ & -- \\
PGD train & 0.1 & $76.61\pm0.26$ & $18.29\pm0.62$ & $8.89\pm4.22$ & $0.84\pm0.09$ \\
PGD train & 0.2 & $75.55\pm0.16$ & $18.48\pm0.82$ & $9.47\pm5.26$ & $1.86\pm0.35$ \\
PGD train & 0.3 & $74.20\pm0.27$ & $18.89\pm0.95$ & $9.87\pm4.99$ & $3.30\pm0.67$ \\
Fair PGD ($\tau{=}0.015$) & 0.0 & $74.32\pm0.94$ & $10.04\pm1.57$ & $-8.70\pm5.38$ & -- \\
Fair PGD ($\tau{=}0.015$) & 0.1 & $73.53\pm1.04$ & $9.86\pm1.54$ & $-8.87\pm5.35$ & $0.69\pm0.12$ \\
Fair PGD ($\tau{=}0.015$) & 0.2 & $72.26\pm0.91$ & $9.61\pm1.35$ & $-9.53\pm5.15$ & $1.83\pm0.06$ \\
Fair PGD ($\tau{=}0.015$) & 0.3 & $71.09\pm0.71$ & $9.59\pm1.13$ & $-9.79\pm4.54$ & $3.03\pm0.08$ \\
\bottomrule
\end{tabular*}
\end{table*}
\FloatBarrier

Table~\ref{tab:app-auc-central} gives the corresponding area under the receiver-operating-characteristic curve (AUC). Under PGD at $\varepsilon=0.3$, AUC is 0.8568 for PGD-trained logistic regression and 0.8702 for the PGD-trained MLP, compared with 0.7580 and 0.7314 under ERM. Fair PGD retains strong ranking performance at the representative budget, whereas static augmentation deteriorates rapidly under model-specific attack. This agreement between thresholded and ranking metrics shows that the central robust comparison is not an artifact of the $0.5$ decision threshold.

\begin{table*}[!htbp]
\centering
\caption{\textbf{Complete AUC record for the central evaluation.} Central methods report three-seed mean $\pm$ sample standard deviation. Static augmentation is aggregate-only.}
\label{tab:app-auc-central}
\footnotesize
\textit{Logistic regression}\par
\begin{tabular*}{\textwidth}{@{\extracolsep{\fill}}l c c c c@{}}
\toprule
Training & $\varepsilon=0$ & 0.1 & 0.2 & 0.3 \\
\midrule
ERM & $0.9040\pm0.0042$ & $0.8643\pm0.0032$ & $0.8156\pm0.0016$ & $0.7580\pm0.0017$ \\
PGD train & $0.8890\pm0.0047$ & $0.8785\pm0.0044$ & $0.8680\pm0.0042$ & $0.8568\pm0.0037$ \\
Fair PGD & $0.8855\pm0.0059$ & $0.8751\pm0.0052$ & $0.8650\pm0.0044$ & $0.8544\pm0.0037$ \\
Static aug. & 0.8282 & 0.6486 & 0.4358 & 0.2459 \\
\bottomrule
\end{tabular*}
\medskip
\textit{Small MLP}\par
\begin{tabular*}{\textwidth}{@{\extracolsep{\fill}}l c c c c@{}}
\toprule
Training & $\varepsilon=0$ & 0.1 & 0.2 & 0.3 \\
\midrule
ERM & $0.9111\pm0.0034$ & $0.8645\pm0.0035$ & $0.8044\pm0.0029$ & $0.7314\pm0.0023$ \\
PGD train & $0.9084\pm0.0040$ & $0.8996\pm0.0040$ & $0.8881\pm0.0038$ & $0.8702\pm0.0031$ \\
Fair PGD & $0.9035\pm0.0048$ & $0.8945\pm0.0051$ & $0.8830\pm0.0056$ & $0.8653\pm0.0062$ \\
Static aug. & 0.9081 & 0.1862 & 0.0363 & 0.0237 \\
\bottomrule
\end{tabular*}
\end{table*}
\FloatBarrier

Together, Tables~\ref{tab:app-epsilon-sweep} and~\ref{tab:app-auc-central} show that the strongest-attack comparison in the main paper summarizes a regular path across the complete tested range rather than an isolated endpoint.

\section{Paired seed-level consistency}
\label{app:seed-level-results}

Table~\ref{tab:app-per-seed} resolves the strongest-attack means into their paired replicates. PGD training improves BA and lowers ASR for every listed seed. Fair PGD at $\tau=0.015$ lowers DPD relative to unconstrained PGD training in every seed and preserves a BA margin over ERM in every seed. The central robust and parity effects are therefore not carried by one favorable split.

\begin{table*}[!htbp]
\centering
\caption{\textbf{Paired seed-level values at PGD $\varepsilon=0.3$.} BA and ASR are percent; DPD and signed EOpp are percentage points.}
\label{tab:app-per-seed}
\footnotesize
\begin{tabular*}{\textwidth}{@{\extracolsep{\fill}}l c c c c c c@{}}
\toprule
Model & Training & Seed & BA & DPD & EOpp & ASR \\
\midrule
Logistic regression & ERM & 0 & 64.65 & 23.09 & +10.50 & 13.64 \\
 &  & 1 & 63.99 & 24.67 & +19.43 & 14.36 \\
 &  & 2 & 63.71 & 23.80 & +12.02 & 15.00 \\
 & PGD train & 0 & 72.13 & 19.28 & +14.21 & 2.08 \\
 &  & 1 & 71.53 & 19.74 & +21.11 & 2.56 \\
 &  & 2 & 71.99 & 18.52 & +12.87 & 2.48 \\
 & Fair PGD & 0 & 67.41 & 8.60 & -6.49 & 1.97 \\
 &  & 1 & 67.39 & 9.91 & +3.05 & 1.91 \\
 &  & 2 & 67.09 & 9.00 & -4.78 & 1.93 \\
\addlinespace[2pt]
Small MLP & ERM & 0 & 63.04 & 23.51 & +3.15 & 18.40 \\
 &  & 1 & 62.65 & 25.80 & +12.26 & 18.58 \\
 &  & 2 & 62.55 & 23.35 & +3.85 & 18.95 \\
 & PGD train & 0 & 74.32 & 18.06 & +6.41 & 2.70 \\
 &  & 1 & 74.38 & 19.92 & +15.59 & 3.17 \\
 &  & 2 & 73.89 & 18.68 & +7.62 & 4.02 \\
 & Fair PGD & 0 & 70.39 & 8.39 & -13.88 & 2.94 \\
 &  & 1 & 71.81 & 10.64 & -4.91 & 3.08 \\
 &  & 2 & 71.06 & 9.74 & -10.58 & 3.08 \\
\addlinespace[2pt]
\bottomrule
\end{tabular*}
\end{table*}
\FloatBarrier

The seed-level EOpp values also preserve the necessary qualification. At the representative budget, two logistic seeds are negative and one remains positive, whereas all three MLP seeds are negative. The complete sweep supports a seed-averaged direction change, not one universal crossing budget.

\section{Complete fairness-budget paths}
\label{app:full-budget-paths}

Table~\ref{tab:app-tolerance-logistic} reports the complete five-budget by four-attack grid for logistic regression. Its rows show a regular seed-averaged exchange: each tighter budget lowers hard DPD and balanced accuracy at every evaluation strength, while signed EOpp moves through zero as the group TPR ordering changes.

\begin{table*}[!htbp]
\centering
\caption{\textbf{Complete fairness-budget grid for logistic regression.} Three-seed mean $\pm$ sample standard deviation under PGD evaluation.}
\label{tab:app-tolerance-logistic}
\footnotesize
\textit{Balanced accuracy (\%)}\par
\begin{tabular*}{\textwidth}{@{\extracolsep{\fill}}l c c c c@{}}
\toprule
$\tau$ & $\varepsilon=0$ & 0.1 & 0.2 & 0.3 \\
\midrule
0.025 & $72.68\pm0.20$ & $72.12\pm0.07$ & $71.41\pm0.06$ & $70.67\pm0.10$ \\
0.020 & $71.29\pm0.20$ & $70.59\pm0.32$ & $69.73\pm0.31$ & $69.19\pm0.55$ \\
0.015 & $69.14\pm0.22$ & $68.55\pm0.27$ & $67.94\pm0.29$ & $67.30\pm0.18$ \\
0.010 & $67.31\pm0.32$ & $66.58\pm0.51$ & $65.94\pm0.73$ & $65.33\pm0.81$ \\
0.005 & $61.12\pm1.62$ & $60.44\pm1.44$ & $59.76\pm1.36$ & $58.98\pm1.33$ \\
\bottomrule
\end{tabular*}
\medskip
\textit{Hard DPD (pp)}\par
\begin{tabular*}{\textwidth}{@{\extracolsep{\fill}}l c c c c@{}}
\toprule
$\tau$ & $\varepsilon=0$ & 0.1 & 0.2 & 0.3 \\
\midrule
0.025 & $14.67\pm1.10$ & $14.75\pm1.26$ & $15.05\pm1.42$ & $15.37\pm1.49$ \\
0.020 & $12.05\pm0.53$ & $12.25\pm0.75$ & $12.02\pm0.70$ & $11.92\pm0.80$ \\
0.015 & $9.13\pm0.44$ & $9.17\pm0.38$ & $9.23\pm0.39$ & $9.17\pm0.67$ \\
0.010 & $6.73\pm0.19$ & $6.64\pm0.10$ & $6.29\pm0.17$ & $6.26\pm0.35$ \\
0.005 & $1.25\pm1.38$ & $1.09\pm1.59$ & $0.64\pm1.64$ & $0.25\pm1.57$ \\
\bottomrule
\end{tabular*}
\medskip
\textit{Signed EOpp (pp)}\par
\begin{tabular*}{\textwidth}{@{\extracolsep{\fill}}l c c c c@{}}
\toprule
$\tau$ & $\varepsilon=0$ & 0.1 & 0.2 & 0.3 \\
\midrule
0.025 & $9.67\pm6.25$ & $9.52\pm6.38$ & $9.52\pm6.54$ & $9.76\pm6.36$ \\
0.020 & $3.58\pm4.91$ & $3.73\pm5.81$ & $3.13\pm4.92$ & $2.95\pm4.98$ \\
0.015 & $-2.97\pm4.37$ & $-3.44\pm4.30$ & $-3.31\pm4.30$ & $-2.74\pm5.09$ \\
0.010 & $-7.38\pm2.46$ & $-7.67\pm2.22$ & $-7.95\pm2.98$ & $-8.12\pm3.17$ \\
0.005 & $-17.88\pm5.10$ & $-18.50\pm5.45$ & $-19.48\pm5.46$ & $-19.87\pm6.16$ \\
\bottomrule
\end{tabular*}
\medskip
\textit{Attack success (\%)}\par
\begin{tabular*}{\textwidth}{@{\extracolsep{\fill}}l c c c c@{}}
\toprule
$\tau$ & $\varepsilon=0$ & 0.1 & 0.2 & 0.3 \\
\midrule
0.025 & $0.00\pm0.00$ & $0.58\pm0.29$ & $1.33\pm0.29$ & $2.14\pm0.26$ \\
0.020 & $0.00\pm0.00$ & $0.69\pm0.17$ & $1.40\pm0.17$ & $1.92\pm0.35$ \\
0.015 & $0.00\pm0.00$ & $0.58\pm0.08$ & $1.24\pm0.11$ & $1.94\pm0.03$ \\
0.010 & $0.00\pm0.00$ & $0.63\pm0.21$ & $1.26\pm0.33$ & $1.87\pm0.35$ \\
0.005 & $0.00\pm0.00$ & $0.58\pm0.25$ & $1.10\pm0.35$ & $1.74\pm0.35$ \\
\bottomrule
\end{tabular*}
\medskip
\end{table*}
\FloatBarrier

Table~\ref{tab:app-tolerance-mlp} gives the corresponding MLP grid. The nonlinear model follows the same mean pattern and retains higher balanced accuracy at the representative interior budget, while its EOpp direction changes earlier along the sampled path. The few local run-level exceptions occur only between neighboring budgets and do not alter the mean ordering.

\begin{table*}[!htbp]
\centering
\caption{\textbf{Complete fairness-budget grid for small MLP.} Three-seed mean $\pm$ sample standard deviation under PGD evaluation.}
\label{tab:app-tolerance-mlp}
\footnotesize
\textit{Balanced accuracy (\%)}\par
\begin{tabular*}{\textwidth}{@{\extracolsep{\fill}}l c c c c@{}}
\toprule
$\tau$ & $\varepsilon=0$ & 0.1 & 0.2 & 0.3 \\
\midrule
0.025 & $76.39\pm0.24$ & $75.54\pm0.17$ & $74.49\pm0.14$ & $73.07\pm0.25$ \\
0.020 & $75.47\pm0.32$ & $74.55\pm0.37$ & $73.33\pm0.34$ & $72.11\pm0.34$ \\
0.015 & $74.32\pm0.94$ & $73.53\pm1.04$ & $72.26\pm0.91$ & $71.09\pm0.71$ \\
0.010 & $72.95\pm0.42$ & $72.00\pm0.40$ & $71.00\pm0.50$ & $69.89\pm0.64$ \\
0.005 & $69.93\pm0.42$ & $69.05\pm0.42$ & $68.15\pm0.32$ & $66.95\pm0.38$ \\
\bottomrule
\end{tabular*}
\medskip
\textit{Hard DPD (pp)}\par
\begin{tabular*}{\textwidth}{@{\extracolsep{\fill}}l c c c c@{}}
\toprule
$\tau$ & $\varepsilon=0$ & 0.1 & 0.2 & 0.3 \\
\midrule
0.025 & $14.97\pm0.40$ & $14.94\pm0.27$ & $14.85\pm0.43$ & $14.81\pm0.34$ \\
0.020 & $12.76\pm0.93$ & $12.64\pm1.01$ & $12.51\pm1.00$ & $12.18\pm0.92$ \\
0.015 & $10.04\pm1.57$ & $9.86\pm1.54$ & $9.61\pm1.35$ & $9.59\pm1.13$ \\
0.010 & $7.85\pm0.79$ & $7.45\pm0.95$ & $7.27\pm1.13$ & $7.27\pm1.27$ \\
0.005 & $4.05\pm0.54$ & $3.62\pm0.73$ & $3.16\pm0.57$ & $2.91\pm0.50$ \\
\bottomrule
\end{tabular*}
\medskip
\textit{Signed EOpp (pp)}\par
\begin{tabular*}{\textwidth}{@{\extracolsep{\fill}}l c c c c@{}}
\toprule
$\tau$ & $\varepsilon=0$ & 0.1 & 0.2 & 0.3 \\
\midrule
0.025 & $2.11\pm4.44$ & $2.12\pm4.48$ & $2.36\pm4.84$ & $1.75\pm4.50$ \\
0.020 & $-2.84\pm4.81$ & $-2.88\pm5.05$ & $-2.72\pm5.06$ & $-3.34\pm5.08$ \\
0.015 & $-8.70\pm5.38$ & $-8.87\pm5.35$ & $-9.53\pm5.15$ & $-9.79\pm4.54$ \\
0.010 & $-13.52\pm2.14$ & $-14.50\pm2.93$ & $-14.69\pm3.01$ & $-14.33\pm2.94$ \\
0.005 & $-21.58\pm2.90$ & $-22.33\pm3.10$ & $-22.88\pm2.51$ & $-22.50\pm2.64$ \\
\bottomrule
\end{tabular*}
\medskip
\textit{Attack success (\%)}\par
\begin{tabular*}{\textwidth}{@{\extracolsep{\fill}}l c c c c@{}}
\toprule
$\tau$ & $\varepsilon=0$ & 0.1 & 0.2 & 0.3 \\
\midrule
0.025 & $0.00\pm0.00$ & $0.77\pm0.10$ & $1.70\pm0.08$ & $3.14\pm0.41$ \\
0.020 & $0.00\pm0.00$ & $0.81\pm0.05$ & $1.90\pm0.09$ & $3.06\pm0.21$ \\
0.015 & $0.00\pm0.00$ & $0.69\pm0.12$ & $1.83\pm0.06$ & $3.03\pm0.08$ \\
0.010 & $0.00\pm0.00$ & $0.76\pm0.06$ & $1.68\pm0.25$ & $2.78\pm0.35$ \\
0.005 & $0.00\pm0.00$ & $0.67\pm0.06$ & $1.43\pm0.07$ & $2.47\pm0.22$ \\
\bottomrule
\end{tabular*}
\medskip
\end{table*}
\FloatBarrier

\begin{figure*}[!htbp]
\centering
\includegraphics[width=\textwidth]{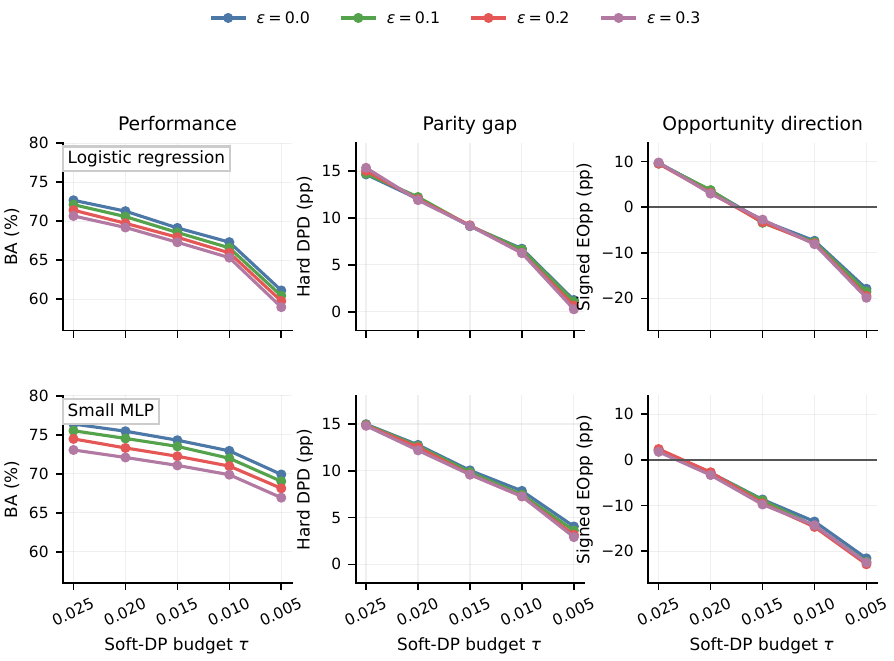}
\caption{\textbf{Complete fairness-budget paths.} Three-seed means for Fair PGD across every budget and PGD evaluation strength. Columns show BA, hard DPD, and signed EOpp; rows separate logistic regression and the small MLP. Lines connect sampled budgets and do not assert global Pareto optimality or a hard-DPD guarantee.}
\label{fig:app-budget-grid}
\end{figure*}

Figure~\ref{fig:app-budget-grid} synthesizes both tables on a common visual scale. The balanced-accuracy and DPD panels make the sampled tradeoff explicit at every PGD strength, while the EOpp panels show that stronger DP control changes a distinct conditional-rate criterion rather than merely shrinking one scalar disparity.

\FloatBarrier

Table~\ref{tab:app-auc-budget} shows that AUC changes regularly along the same budget path. Tighter constraints gradually lower AUC, but the constrained models retain strong ranking performance throughout the tested range. The sampled operating path is therefore visible both in hard-decision metrics and in a threshold-independent ranking metric.

\begin{table*}[!htbp]
\centering
\caption{\textbf{AUC along the complete fairness-budget path.} Three-seed mean $\pm$ sample standard deviation.}
\label{tab:app-auc-budget}
\footnotesize
\textit{Logistic regression}\par
\begin{tabular*}{\textwidth}{@{\extracolsep{\fill}}l c c c c@{}}
\toprule
$\tau$ & $\varepsilon=0$ & 0.1 & 0.2 & 0.3 \\
\midrule
0.025 & $0.8887\pm0.0050$ & $0.8785\pm0.0046$ & $0.8684\pm0.0042$ & $0.8577\pm0.0039$ \\
0.020 & $0.8876\pm0.0053$ & $0.8775\pm0.0049$ & $0.8676\pm0.0044$ & $0.8571\pm0.0040$ \\
0.015 & $0.8855\pm0.0059$ & $0.8751\pm0.0052$ & $0.8650\pm0.0044$ & $0.8544\pm0.0037$ \\
0.010 & $0.8816\pm0.0066$ & $0.8709\pm0.0059$ & $0.8605\pm0.0052$ & $0.8496\pm0.0048$ \\
0.005 & $0.8719\pm0.0087$ & $0.8612\pm0.0080$ & $0.8509\pm0.0072$ & $0.8401\pm0.0066$ \\
\bottomrule
\end{tabular*}
\medskip
\textit{Small MLP}\par
\begin{tabular*}{\textwidth}{@{\extracolsep{\fill}}l c c c c@{}}
\toprule
$\tau$ & $\varepsilon=0$ & 0.1 & 0.2 & 0.3 \\
\midrule
0.025 & $0.9072\pm0.0049$ & $0.8984\pm0.0049$ & $0.8870\pm0.0050$ & $0.8689\pm0.0060$ \\
0.020 & $0.9064\pm0.0046$ & $0.8973\pm0.0047$ & $0.8855\pm0.0050$ & $0.8672\pm0.0051$ \\
0.015 & $0.9035\pm0.0048$ & $0.8945\pm0.0051$ & $0.8830\pm0.0056$ & $0.8653\pm0.0062$ \\
0.010 & $0.8986\pm0.0066$ & $0.8899\pm0.0062$ & $0.8782\pm0.0063$ & $0.8608\pm0.0056$ \\
0.005 & $0.8926\pm0.0069$ & $0.8832\pm0.0075$ & $0.8711\pm0.0079$ & $0.8527\pm0.0085$ \\
\bottomrule
\end{tabular*}
\end{table*}
\FloatBarrier

The complete grid gives $\tau$ a precise empirical interpretation: it is an explicit soft-score operating control from which an externally chosen performance floor and disparity ceiling can select among tested points. It is not a proof that $\sqrt\tau$ bounds hard DPD.

\section{Group-level mechanism and sign reversal}
\label{app:group-results}

Table~\ref{tab:app-group-logistic} decomposes the logistic results into group PPR, TPR, balanced accuracy, and attack success. The table shows that Fair PGD closes DPD primarily by reducing Group~0's initially higher positive-prediction rate, while Group~1 increases more modestly; the TPR columns reveal the same asymmetric movement that later reverses signed EOpp.

\begin{table*}[!htbp]
\centering
\caption{\textbf{Complete group-level metrics for logistic regression.} Three-seed mean $\pm$ sample standard deviation. Group 0 is Male and Group 1 is Female.}
\label{tab:app-group-logistic}
\footnotesize
\textit{Selection and opportunity rates}\par
\begin{tabular*}{\textwidth}{@{\extracolsep{\fill}}l c c c c c@{}}
\toprule
Training & $\varepsilon$ & PPR$_0$ & PPR$_1$ & TPR$_0$ & TPR$_1$ \\
\midrule
ERM & 0.0 & $26.20\pm0.38$ & $7.26\pm0.36$ & $61.62\pm0.92$ & $49.41\pm3.88$ \\
ERM & 0.1 & $27.79\pm0.40$ & $7.74\pm0.45$ & $57.71\pm0.74$ & $45.91\pm3.93$ \\
ERM & 0.2 & $30.35\pm0.35$ & $8.31\pm0.50$ & $53.55\pm1.12$ & $39.46\pm4.74$ \\
ERM & 0.3 & $33.71\pm0.43$ & $9.86\pm0.42$ & $49.22\pm0.80$ & $35.23\pm4.17$ \\
PGD train & 0.0 & $24.86\pm0.44$ & $6.38\pm0.44$ & $57.55\pm0.60$ & $40.22\pm3.68$ \\
PGD train & 0.1 & $25.11\pm0.45$ & $6.49\pm0.46$ & $56.58\pm0.63$ & $39.65\pm3.91$ \\
PGD train & 0.2 & $25.39\pm0.49$ & $6.66\pm0.48$ & $55.62\pm0.53$ & $39.26\pm4.07$ \\
PGD train & 0.3 & $25.88\pm0.53$ & $6.70\pm0.50$ & $54.93\pm0.64$ & $38.87\pm4.23$ \\
Fair PGD ($\tau{=}0.015$) & 0.0 & $16.47\pm0.27$ & $7.33\pm0.46$ & $41.79\pm0.88$ & $44.77\pm3.71$ \\
Fair PGD ($\tau{=}0.015$) & 0.1 & $16.57\pm0.28$ & $7.40\pm0.41$ & $40.93\pm0.95$ & $44.38\pm3.57$ \\
Fair PGD ($\tau{=}0.015$) & 0.2 & $16.76\pm0.25$ & $7.53\pm0.48$ & $40.20\pm0.84$ & $43.51\pm3.75$ \\
Fair PGD ($\tau{=}0.015$) & 0.3 & $16.93\pm0.22$ & $7.76\pm0.65$ & $39.51\pm0.86$ & $42.25\pm4.37$ \\
\bottomrule
\end{tabular*}
\medskip
\textit{Predictive performance and attack response}\par
\begin{tabular*}{\textwidth}{@{\extracolsep{\fill}}l c c c c c@{}}
\toprule
Training & $\varepsilon$ & BA$_0$ & BA$_1$ & ASR$_0$ & ASR$_1$ \\
\midrule
ERM & 0.0 & $75.80\pm0.31$ & $73.80\pm1.97$ & -- & -- \\
ERM & 0.1 & $71.79\pm0.19$ & $71.55\pm1.89$ & $4.98\pm0.19$ & $1.39\pm0.20$ \\
ERM & 0.2 & $66.90\pm0.61$ & $67.59\pm2.33$ & $11.36\pm0.55$ & $3.59\pm0.40$ \\
ERM & 0.3 & $61.29\pm0.56$ & $64.33\pm2.09$ & $18.85\pm0.69$ & $6.31\pm0.59$ \\
PGD train & 0.0 & $73.81\pm0.39$ & $69.11\pm1.79$ & -- & -- \\
PGD train & 0.1 & $72.93\pm0.27$ & $68.72\pm1.90$ & $1.07\pm0.23$ & $0.26\pm0.04$ \\
PGD train & 0.2 & $72.02\pm0.04$ & $68.40\pm1.98$ & $2.18\pm0.36$ & $0.55\pm0.04$ \\
PGD train & 0.3 & $71.16\pm0.06$ & $68.16\pm2.05$ & $3.33\pm0.35$ & $0.69\pm0.06$ \\
Fair PGD ($\tau{=}0.015$) & 0.0 & $68.45\pm0.45$ & $71.14\pm1.82$ & -- & -- \\
Fair PGD ($\tau{=}0.015$) & 0.1 & $67.75\pm0.52$ & $70.88\pm1.76$ & $0.82\pm0.11$ & $0.17\pm0.04$ \\
Fair PGD ($\tau{=}0.015$) & 0.2 & $67.07\pm0.48$ & $70.31\pm1.81$ & $1.65\pm0.14$ & $0.52\pm0.08$ \\
Fair PGD ($\tau{=}0.015$) & 0.3 & $66.45\pm0.45$ & $69.47\pm2.07$ & $2.41\pm0.04$ & $1.10\pm0.12$ \\
\bottomrule
\end{tabular*}
\end{table*}
\FloatBarrier

Table~\ref{tab:app-group-mlp} reports the same quantities for the MLP. The nonlinear model reproduces the mechanism: Group~0 PPR and TPR fall as the budget tightens, Group~1 rises by a smaller amount, and group-specific performance and attack-success rates make the distribution of gains and costs visible.

\begin{table*}[!htbp]
\centering
\caption{\textbf{Complete group-level metrics for small MLP.} Three-seed mean $\pm$ sample standard deviation. Group 0 is Male and Group 1 is Female.}
\label{tab:app-group-mlp}
\footnotesize
\textit{Selection and opportunity rates}\par
\begin{tabular*}{\textwidth}{@{\extracolsep{\fill}}l c c c c c@{}}
\toprule
Training & $\varepsilon$ & PPR$_0$ & PPR$_1$ & TPR$_0$ & TPR$_1$ \\
\midrule
ERM & 0.0 & $27.56\pm0.22$ & $8.92\pm0.79$ & $65.16\pm0.95$ & $57.17\pm3.69$ \\
ERM & 0.1 & $30.48\pm0.29$ & $9.89\pm0.90$ & $60.68\pm0.86$ & $52.60\pm4.37$ \\
ERM & 0.2 & $34.45\pm0.44$ & $11.31\pm1.16$ & $55.92\pm0.88$ & $48.42\pm5.19$ \\
ERM & 0.3 & $38.07\pm0.36$ & $13.85\pm1.43$ & $50.25\pm0.63$ & $43.83\pm4.93$ \\
PGD train & 0.0 & $26.74\pm0.12$ & $8.44\pm0.70$ & $63.45\pm1.01$ & $54.72\pm3.94$ \\
PGD train & 0.1 & $26.79\pm0.21$ & $8.51\pm0.64$ & $62.16\pm0.73$ & $53.26\pm3.50$ \\
PGD train & 0.2 & $27.00\pm0.13$ & $8.52\pm0.76$ & $60.79\pm0.94$ & $51.32\pm4.32$ \\
PGD train & 0.3 & $27.57\pm0.25$ & $8.68\pm0.73$ & $59.25\pm0.96$ & $49.38\pm4.07$ \\
Fair PGD ($\tau{=}0.015$) & 0.0 & $20.30\pm1.09$ & $10.26\pm0.48$ & $52.25\pm2.83$ & $60.95\pm2.58$ \\
Fair PGD ($\tau{=}0.015$) & 0.1 & $20.23\pm1.03$ & $10.37\pm0.51$ & $51.02\pm2.91$ & $59.88\pm2.56$ \\
Fair PGD ($\tau{=}0.015$) & 0.2 & $20.20\pm0.90$ & $10.59\pm0.47$ & $49.07\pm2.63$ & $58.60\pm2.53$ \\
Fair PGD ($\tau{=}0.015$) & 0.3 & $20.47\pm0.76$ & $10.88\pm0.40$ & $47.55\pm2.17$ & $57.33\pm2.40$ \\
\bottomrule
\end{tabular*}
\medskip
\textit{Predictive performance and attack response}\par
\begin{tabular*}{\textwidth}{@{\extracolsep{\fill}}l c c c c c@{}}
\toprule
Training & $\varepsilon$ & BA$_0$ & BA$_1$ & ASR$_0$ & ASR$_1$ \\
\midrule
ERM & 0.0 & $77.39\pm0.53$ & $77.24\pm1.59$ & -- & -- \\
ERM & 0.1 & $72.00\pm0.54$ & $74.12\pm1.90$ & $6.99\pm0.14$ & $2.17\pm0.02$ \\
ERM & 0.2 & $65.64\pm0.41$ & $70.95\pm2.21$ & $15.48\pm0.04$ & $4.73\pm0.30$ \\
ERM & 0.3 & $58.87\pm0.23$ & $66.93\pm1.93$ & $24.24\pm0.27$ & $8.61\pm0.44$ \\
PGD train & 0.0 & $76.74\pm0.69$ & $76.13\pm1.80$ & -- & -- \\
PGD train & 0.1 & $75.76\pm0.52$ & $75.27\pm1.57$ & $1.06\pm0.16$ & $0.43\pm0.09$ \\
PGD train & 0.2 & $74.61\pm0.55$ & $74.16\pm1.96$ & $2.37\pm0.54$ & $0.94\pm0.01$ \\
PGD train & 0.3 & $73.07\pm0.53$ & $72.97\pm1.82$ & $4.24\pm0.92$ & $1.59\pm0.20$ \\
Fair PGD ($\tau{=}0.015$) & 0.0 & $73.27\pm1.22$ & $78.62\pm1.12$ & -- & -- \\
Fair PGD ($\tau{=}0.015$) & 0.1 & $72.43\pm1.32$ & $77.96\pm1.08$ & $0.86\pm0.15$ & $0.39\pm0.07$ \\
Fair PGD ($\tau{=}0.015$) & 0.2 & $71.03\pm1.21$ & $77.11\pm1.12$ & $2.33\pm0.12$ & $0.94\pm0.04$ \\
Fair PGD ($\tau{=}0.015$) & 0.3 & $69.72\pm0.99$ & $76.23\pm1.10$ & $3.85\pm0.12$ & $1.57\pm0.13$ \\
\bottomrule
\end{tabular*}
\end{table*}
\FloatBarrier

\begin{figure*}[!htbp]
\centering
\includegraphics[width=\textwidth]{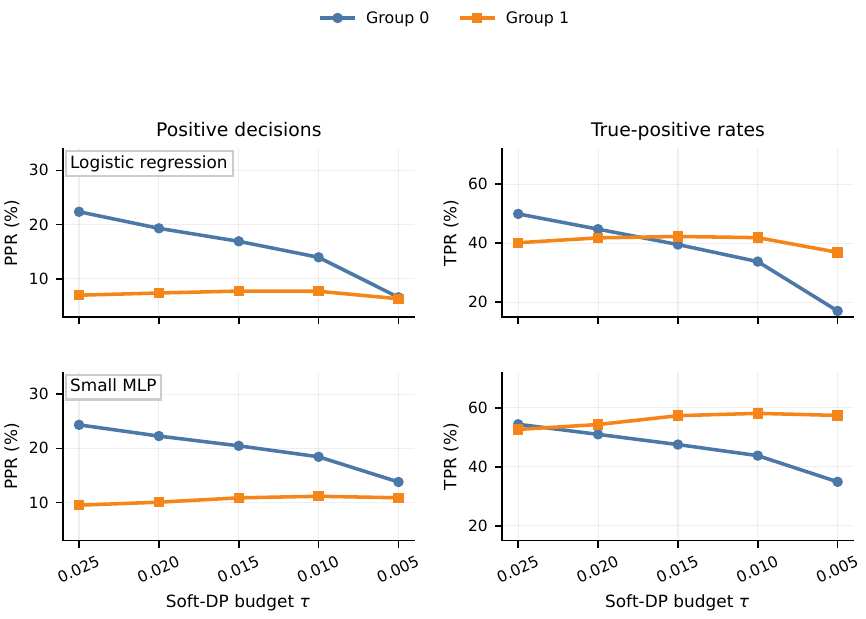}
\caption{\textbf{Group-rate paths under the fairness budget.} Three-seed mean PPR and TPR at PGD $\varepsilon=0.3$ as $\tau$ tightens. Group~0 is Male and Group~1 is Female. The figure exposes the rate redistribution behind the aggregate DPD and EOpp values.}
\label{fig:app-group-rates}
\end{figure*}

Figure~\ref{fig:app-group-rates} places the PPR and TPR paths from both tables side by side at $\varepsilon=0.3$. The figure makes the mechanism visually immediate: the aggregate DPD reduction is generated by unequal rate shifts, not by a uniform improvement shared by both groups.

\FloatBarrier

The measured TPR changes satisfy Equation~\eqref{eq:app-reversal}. For logistic regression, $\Delta_1-\Delta_0=20.30$ percentage points exceeds the initial $e=17.32$ points, producing final EOpp of $-2.97$ points. For the MLP, the relative shift is $17.43$ points against an initial $8.74$-point gap, producing $-8.70$ points. The sign reversal follows exactly from the observed rate redistribution, not from DPD alone. The group BA and ASR columns further show why lower DPD is not automatically a welfare conclusion: the constraint changes who receives positive predictions and how errors are distributed.

\section{FGSM and PGD comparison}
\label{app:attack-comparison}

The principal evaluation uses PGD, but the ERM vulnerability is not unique to that construction. Table~\ref{tab:app-fgsm} gives the exact BA, DPD, and ASR values under the fast gradient sign method (FGSM) and PGD at every nonzero strength. Logistic results agree at the reported precision, and the MLP differences remain small.

\begin{table*}[!htbp]
\centering
\caption{\textbf{FGSM and PGD evaluation of ERM models.} Three-seed mean $\pm$ sample standard deviation. BA and ASR are percent; DPD is percentage points.}
\label{tab:app-fgsm}
\footnotesize
\begin{tabular*}{\textwidth}{@{\extracolsep{\fill}}l c c c c c@{}}
\toprule
Model & $\varepsilon$ & Attack & BA & DPD & ASR \\
\midrule
Logistic regression & 0.1 & FGSM & $73.08\pm0.12$ & $20.05\pm0.84$ & $3.68\pm0.10$ \\
 & 0.1 & PGD & $73.08\pm0.12$ & $20.05\pm0.84$ & $3.68\pm0.10$ \\
 & 0.2 & FGSM & $68.82\pm0.42$ & $22.04\pm0.85$ & $8.56\pm0.51$ \\
 & 0.2 & PGD & $68.82\pm0.42$ & $22.04\pm0.85$ & $8.56\pm0.51$ \\
 & 0.3 & FGSM & $64.12\pm0.48$ & $23.85\pm0.79$ & $14.33\pm0.68$ \\
 & 0.3 & PGD & $64.12\pm0.48$ & $23.85\pm0.79$ & $14.33\pm0.68$ \\
\addlinespace[2pt]
Small MLP & 0.1 & FGSM & $73.78\pm0.23$ & $20.59\pm0.96$ & $5.25\pm0.06$ \\
 & 0.1 & PGD & $73.77\pm0.23$ & $20.59\pm0.97$ & $5.26\pm0.06$ \\
 & 0.2 & FGSM & $68.59\pm0.08$ & $23.10\pm1.36$ & $11.61\pm0.15$ \\
 & 0.2 & PGD & $68.58\pm0.11$ & $23.14\pm1.36$ & $11.63\pm0.16$ \\
 & 0.3 & FGSM & $62.77\pm0.24$ & $24.14\pm1.34$ & $18.58\pm0.28$ \\
 & 0.3 & PGD & $62.75\pm0.26$ & $24.22\pm1.37$ & $18.64\pm0.28$ \\
\addlinespace[2pt]
\bottomrule
\end{tabular*}
\end{table*}
\FloatBarrier

\begin{figure*}[!htbp]
\centering
\includegraphics[width=\textwidth]{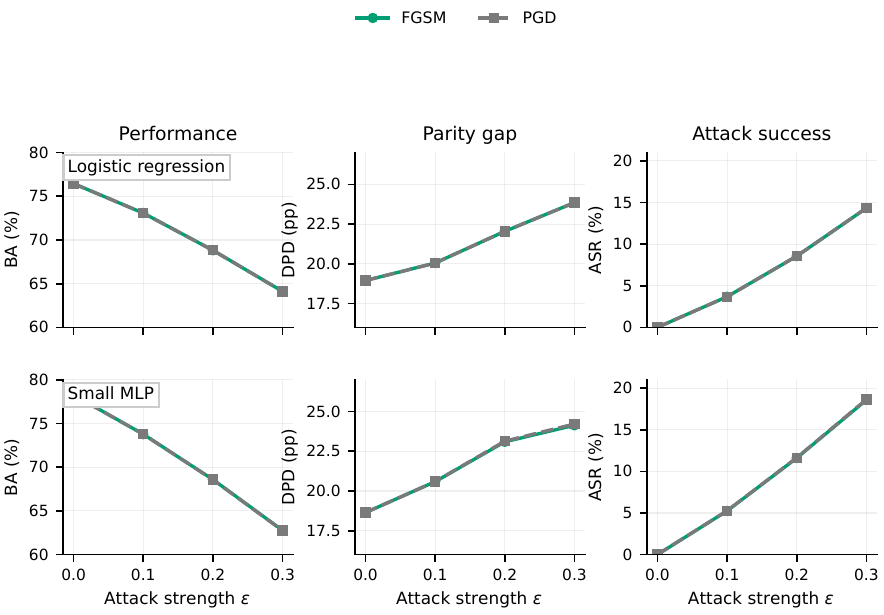}
\caption{\textbf{FGSM and PGD evaluation of ERM models.} The panels compare BA, DPD, and ASR over the tested strengths for both model classes. PGD remains the principal iterative white-box evaluation; close agreement does not constitute certification or prove attack optimality.}
\label{fig:app-attack-comparison}
\end{figure*}

Figure~\ref{fig:app-attack-comparison} visualizes the same comparison across both model classes. The near-overlapping trajectories show that the baseline degradation is not created by one attack implementation, while PGD remains the principal iterative white-box evaluation.

\FloatBarrier

Table~\ref{tab:app-auc-attacks} confirms the same conclusion with a threshold-independent ranking metric. The close AUC agreement strengthens the attack comparison without implying certification or attack optimality.

\begin{table*}[!htbp]
\centering
\caption{\textbf{AUC under FGSM and PGD evaluation of ERM models.} Three-seed mean $\pm$ sample standard deviation.}
\label{tab:app-auc-attacks}
\footnotesize
\begin{tabular*}{\textwidth}{@{\extracolsep{\fill}}l c c c c c@{}}
\toprule
Model & Attack & $\varepsilon=0$ & 0.1 & 0.2 & 0.3 \\
\midrule
Logistic regression & FGSM & $0.9040\pm0.0042$ & $0.8643\pm0.0032$ & $0.8156\pm0.0016$ & $0.7580\pm0.0017$ \\
 & PGD & $0.9040\pm0.0042$ & $0.8643\pm0.0032$ & $0.8156\pm0.0016$ & $0.7580\pm0.0017$ \\
\addlinespace[2pt]
Small MLP & FGSM & $0.9111\pm0.0034$ & $0.8645\pm0.0035$ & $0.8046\pm0.0028$ & $0.7321\pm0.0022$ \\
 & PGD & $0.9111\pm0.0034$ & $0.8645\pm0.0035$ & $0.8044\pm0.0029$ & $0.7314\pm0.0023$ \\
\addlinespace[2pt]
\bottomrule
\end{tabular*}
\end{table*}
\FloatBarrier

The agreement in Tables~\ref{tab:app-fgsm} and~\ref{tab:app-auc-attacks} and Figure~\ref{fig:app-attack-comparison} supports the use of PGD as the primary evaluation without enlarging the claim. The study remains a single-start continuous feature-space evaluation.

\section{Static adversarial augmentation as a negative control}
\label{app:static-control}

Table~\ref{tab:app-static} compares on-the-fly PGD training with a fixed adversarially augmented dataset. The static model is trained on clean data concatenated with ten PGD copies generated by a frozen ERM model. Under attacks generated for the resulting trained model, static BA collapses and ASR rises sharply, whereas on-the-fly PGD training remains stable. The control therefore shows why model-current attack generation is the relevant robust-training comparator in this study.

\begin{table*}[!htbp]
\centering
\caption{\textbf{Static adversarial augmentation is a negative control.} Static rows are aggregate means; seed-level evaluation values were not retained. Fairness columns are excluded because the retained static fairness views did not pass the cross-file audit.}
\label{tab:app-static}
\footnotesize
\begin{tabular*}{\textwidth}{@{\extracolsep{\fill}}l c c c c c@{}}
\toprule
Model & $\varepsilon$ & BA static & BA online & ASR static & ASR online \\
\midrule
Logistic regression & 0.0 & 69.56 & $74.02\pm0.19$ & -- & -- \\
 & 0.1 & 54.98 & $73.28\pm0.11$ & 15.44 & $0.77\pm0.16$ \\
 & 0.2 & 40.38 & $72.54\pm0.29$ & 32.20 & $1.59\pm0.25$ \\
 & 0.3 & 29.58 & $71.88\pm0.31$ & 46.80 & $2.37\pm0.26$ \\
\addlinespace[2pt]
Small MLP & 0.0 & 78.61 & $77.53\pm0.39$ & -- & -- \\
 & 0.1 & 34.12 & $76.61\pm0.26$ & 45.96 & $0.84\pm0.09$ \\
 & 0.2 & 13.17 & $75.55\pm0.16$ & 78.77 & $1.86\pm0.35$ \\
 & 0.3 & 4.45 & $74.20\pm0.27$ & 93.61 & $3.30\pm0.67$ \\
\addlinespace[2pt]
\bottomrule
\end{tabular*}
\end{table*}
\FloatBarrier

The static rows are aggregate means because seed-level evaluation values were not retained; no paired inference is made. The retained static fairness views also fail the cross-file consistency audit, so Table~\ref{tab:app-static} reports only verified predictive and attack metrics. This is a transparent failure case, not a source of a fairness claim. The feasibility and epoch summaries reported above contain the relevant training-history evidence.

\setcounter{section}{11}
\section{Reproducibility and artifact map}
\label{app:artifact-map}

The numerical results in the tables and empirical figures were generated from frozen result files. Table~\ref{tab:app-artifact-map} records the original analysis artifacts and their SHA-256 prefixes; its source-file hashes identify that analysis snapshot rather than the present manuscript. Separate CSV files supplied aggregate and group metrics for the 13 appendix tables and three vector figures. No numerical value was inferred from a rendered plot.

\begin{table*}[!htbp]
\centering
\caption{\textbf{Original analysis artifact record.} File names and SHA-256 prefixes identify artifacts recorded during the analysis. The source-file hashes refer to that analysis snapshot, not to this revised manuscript. The table is a provenance record, not a list of files supplied with the typesetting source.}
\label{tab:app-artifact-map}
\footnotesize
\begin{tabular*}{\textwidth}{@{\extracolsep{\fill}}lll@{}}
\toprule
Artifact & File & SHA-256 prefix \\
\midrule
Main source & \texttt{main.tex} & \texttt{b1b62a6ab2} \\
Appendix source & \texttt{appendix\_content.tex} & \texttt{e15441ace7} \\
Bibliography & \texttt{references.bib} & \texttt{11e8e45556} \\
Appendix generator & \texttt{generate\_appendix\_assets.py} & \texttt{2f4f0393f4} \\
ERM PGD results & \texttt{pgd\_combined.csv} & \texttt{816adfa334} \\
PGD-training results & \texttt{pgd\_train\_combined.csv} & \texttt{e8ac869198} \\
Fairness-budget results & \texttt{fair\_tau\_combined.csv} & \texttt{6485114a54} \\
Fair group 0 & \texttt{fair\_tau\_group\_0.csv} & \texttt{00f7b8c0f0} \\
Fair group 1 & \texttt{fair\_tau\_group\_1.csv} & \texttt{d554aaf618} \\
Static control & \texttt{static\_pgd\_combined.csv} & \texttt{1a80132136} \\
\bottomrule
\end{tabular*}
\end{table*}
\FloatBarrier

The typesetting source contains the manuscript, bibliography, style file, tables, and figure assets needed to reproduce this document. It does not bundle the experiment code, result CSV files, or analysis generator listed in Table~\ref{tab:app-artifact-map}. To compile the paper, run
\begin{verbatim}
pdflatex main.tex
bibtex main
pdflatex main.tex
pdflatex main.tex
\end{verbatim}
The protocol and complete result tables document the experimental record; rebuilding the PDF is distinct from regenerating the experiments.
\clearpage
\section*{NeurIPS Paper Checklist}

\begin{enumerate}
\item {\bf Claims}
    \item[] Question: Do the main claims made in the abstract and introduction accurately reflect the paper's contributions and scope?
    \item[] Answer: \answerYes{}
    \item[] Justification: The abstract and introduction state the budgeted formulation, controlled Adult setting, two model classes, three paired seeds, four evaluation strengths, five budgets, principal numerical results, and the exact scope of the EOpp theorem. Section~\ref{sec:discussion} limits generalization to this dataset, model set, and feature-space threat model.

\item {\bf Limitations}
    \item[] Question: Does the paper discuss the limitations of the work performed by the authors?
    \item[] Answer: \answerYes{}
    \item[] Justification: Section~\ref{sec:discussion} states the one-dataset, binary-group, two-model, three-seed, single-random-start, feature-space-attack, surrogate-to-hard-metric, and missing-comparator limits. It also discusses the leveling-down component and the absence of semantic or certified robustness.

\item {\bf Theory assumptions and proofs}
    \item[] Question: For each theoretical result, does the paper provide the full set of assumptions and a complete (and correct) proof?
    \item[] Answer: \answerYes{}
    \item[] Justification: Theorem~\ref{thm:dp-eopp-direction} states its support assumptions and sign convention, and Section~\ref{sec:method} contains the complete proof. The empirical application is identified separately in Section~\ref{sec:results}.

\item {\bf Experimental result reproducibility}
    \item[] Question: Does the paper fully disclose all the information needed to reproduce the main experimental results of the paper to the extent that it affects the main claims and/or conclusions of the paper (regardless of whether the code and data are provided or not)?
    \item[] Answer: \answerYes{}
    \item[] Justification: Sections~\ref{sec:method} and~\ref{sec:experimental-setup} specify the data, splits, preprocessing, models, attacks, budgets, optimizer, feasibility updates, checkpoint rule, metrics, and paired-replicate reporting needed to reconstruct the central experiments.
    \item[] Guidelines:
    \begin{itemize}
        \item The answer \answerNA{} means that the paper does not include experiments.
        \item If the paper includes experiments, a \answerNo{} answer to this question will not be perceived well by the reviewers: Making the paper reproducible is important, regardless of whether the code and data are provided or not.
        \item If the contribution is a dataset and\slash or model, the authors should describe the steps taken to make their results reproducible or verifiable.
        \item Depending on the contribution, reproducibility can be accomplished in various ways. For example, if the contribution is a novel architecture, describing the architecture fully might suffice, or if the contribution is a specific model and empirical evaluation, it may be necessary to either make it possible for others to replicate the model with the same dataset, or provide access to the model. In general, releasing code and data is often one good way to accomplish this, but reproducibility can also be provided via detailed instructions for how to replicate the results, access to a hosted model (e.g., in the case of a large language model), releasing of a model checkpoint, or other means that are appropriate to the research performed.
        \item While NeurIPS does not require releasing code, the conference does require all submissions to provide some reasonable avenue for reproducibility, which may depend on the nature of the contribution. For example
        \begin{enumerate}
            \item If the contribution is primarily a new algorithm, the paper should make it clear how to reproduce that algorithm.
            \item If the contribution is primarily a new model architecture, the paper should describe the architecture clearly and fully.
            \item If the contribution is a new model (e.g., a large language model), then there should either be a way to access this model for reproducing the results or a way to reproduce the model (e.g., with an open-source dataset or instructions for how to construct the dataset).
            \item We recognize that reproducibility may be tricky in some cases, in which case authors are welcome to describe the particular way they provide for reproducibility. In the case of closed-source models, it may be that access to the model is limited in some way (e.g., to registered users), but it should be possible for other researchers to have some path to reproducing or verifying the results.
        \end{enumerate}
    \end{itemize}

\item {\bf Open access to data and code}
    \item[] Question: Does the paper provide open access to the data and code, with sufficient instructions to faithfully reproduce the main experimental results, as described in supplemental material?
    \item[] Answer: \answerNo{}
    \item[] Justification: Adult is openly available through UCI/OpenML. This paper provides the experimental protocol and complete result tables but does not provide a public research-code release.

\item {\bf Experimental setting/details}
    \item[] Question: Does the paper specify all the training and test details (e.g., data splits, hyperparameters, how they were chosen, type of optimizer) necessary to understand the results?
    \item[] Answer: \answerYes{}
    \item[] Justification: Section~\ref{sec:experimental-setup} gives the paired splits, preprocessing, feature mask, models, optimizer, training and evaluation PGD schedules, tolerance sweep, checkpoint criteria, metrics, and reporting units.

\item {\bf Experiment statistical significance}
    \item[] Question: Does the paper report error bars suitably and correctly defined or other appropriate information about the statistical significance of the experiments?
    \item[] Answer: \answerYes{}
    \item[] Justification: Figures~\ref{fig:robustness-disparity}--\ref{fig:budget-frontier-eopp} and Tables~\ref{tab:key-results}--\ref{tab:group-mechanism} report three-replicate means with individual seeds or sample standard deviations. Section~\ref{sec:experimental-setup} states what varies across replicates; no normality, confidence-interval, or $p$-value claim is made.
    \item[] Guidelines:
    \begin{itemize}
        \item The answer \answerNA{} means that the paper does not include experiments.
        \item The authors should answer \answerYes{} if the results are accompanied by error bars, confidence intervals, or statistical significance tests, at least for the experiments that support the main claims of the paper.
        \item The factors of variability that the error bars are capturing should be clearly stated (for example, train/test split, initialization, random drawing of some parameter, or overall run with given experimental conditions).
        \item The method for calculating the error bars should be explained (closed form formula, call to a library function, bootstrap, etc.)
        \item The assumptions made should be given (e.g., Normally distributed errors).
        \item It should be clear whether the error bar is the standard deviation or the standard error of the mean.
        \item It is OK to report 1-sigma error bars, but one should state it. The authors should preferably report a 2-sigma error bar than state that they have a 96\% CI, if the hypothesis of Normality of errors is not verified.
        \item For asymmetric distributions, the authors should be careful not to show in tables or figures symmetric error bars that would yield results that are out of range (e.g. negative error rates).
        \item If error bars are reported in tables or plots, the authors should explain in the text how they were calculated and reference the corresponding figures or tables in the text.
    \end{itemize}

\item {\bf Experiments compute resources}
    \item[] Question: For each experiment, does the paper provide sufficient information on the computer resources (type of compute workers, memory, time of execution) needed to reproduce the experiments?
    \item[] Answer: \answerNo{}
    \item[] Justification: The implementation is CPU-compatible and the model and data scale are disclosed, but exact per-run wall time, peak memory, and total project compute were not retained. The paper makes no efficiency or scaling claim.

\item {\bf Code of ethics}
    \item[] Question: Does the research conducted in the paper conform, in every respect, with the NeurIPS Code of Ethics \url{https://neurips.cc/public/EthicsGuidelines}?
    \item[] Answer: \answerYes{}
    \item[] Justification: The submitting authors have read the NeurIPS Code of Ethics and confirm that the research conforms to it. The study uses a public benchmark, introduces no human-subject interaction or high-risk model release, and Section~\ref{sec:discussion} addresses discrimination, proxy structure, leveling down, and deployment limits.
    \item[] Guidelines:
    \begin{itemize}
        \item The answer \answerNA{} means that the authors have not reviewed the NeurIPS Code of Ethics.
        \item If the authors answer \answerNo{}, they should explain the special circumstances that require a deviation from the Code of Ethics.
        \item The authors should make sure to preserve anonymity (e.g., if there is a special consideration due to laws or regulations in their jurisdiction).
    \end{itemize}

\item {\bf Broader impacts}
    \item[] Question: Does the paper discuss both potential positive societal impacts and negative societal impacts of the work performed?
    \item[] Answer: \answerYes{}
    \item[] Justification: Section~\ref{sec:discussion} discusses the constructive value of auditable joint robustness/fairness optimization and the risks of proxy discrimination, metric substitution, leveling down, and uncritical deployment.
    \item[] Guidelines:
    \begin{itemize}
        \item The answer \answerNA{} means that there is no societal impact of the work performed.
        \item If the authors answer \answerNA{} or \answerNo{}, they should explain why their work has no societal impact or why the paper does not address societal impact.
        \item Examples of negative societal impacts include potential malicious or unintended uses (e.g., disinformation, generating fake profiles, surveillance), fairness considerations (e.g., deployment of technologies that could make decisions that unfairly impact specific groups), privacy considerations, and security considerations.
        \item The conference expects that many papers will be foundational research and not tied to particular applications, let alone deployments. However, if there is a direct path to any negative applications, the authors should point it out. For example, it is legitimate to point out that an improvement in the quality of generative models could be used to generate Deepfakes for disinformation. On the other hand, it is not needed to point out that a generic algorithm for optimizing neural networks could enable people to train models that generate Deepfakes faster.
        \item The authors should consider possible harms that could arise when the technology is being used as intended and functioning correctly, harms that could arise when the technology is being used as intended but gives incorrect results, and harms following from (intentional or unintentional) misuse of the technology.
        \item If there are negative societal impacts, the authors could also discuss possible mitigation strategies (e.g., gated release of models, providing defenses in addition to attacks, mechanisms for monitoring misuse, mechanisms to monitor how a system learns from feedback over time, improving the efficiency and accessibility of ML).
    \end{itemize}

\item {\bf Safeguards}
    \item[] Question: Does the paper describe safeguards that have been put in place for responsible release of data or models that have a high risk for misuse (e.g., pre-trained language models, image generators, or scraped datasets)?
    \item[] Answer: \answerNA{}
    \item[] Justification: The work releases no high-risk pretrained model, scraped corpus, or dual-use system that requires controlled access or safety filtering.

\item {\bf Licenses for existing assets}
    \item[] Question: Are the creators or original owners of assets (e.g., code, data, models), used in the paper, properly credited and are the license and terms of use explicitly mentioned and properly respected?
    \item[] Answer: \answerYes{}
    \item[] Justification: The Adult benchmark is cited by DOI, the study uses OpenML data identifier 1590, and Adult is distributed by the UCI Machine Learning Repository under CC BY 4.0. No third-party model or private dataset is redistributed.

\item {\bf New assets}
    \item[] Question: Are new assets introduced in the paper well documented and is the documentation provided alongside the assets?
    \item[] Answer: \answerNA{}
    \item[] Justification: The paper releases no new dataset, pretrained model, or research-code archive. The LaTeX source supports document reproduction rather than constituting a new research asset.

\item {\bf Crowdsourcing and research with human subjects}
    \item[] Question: For crowdsourcing experiments and research with human subjects, does the paper include the full text of instructions given to participants and screenshots, if applicable, as well as details about compensation (if any)?
    \item[] Answer: \answerNA{}
    \item[] Justification: The study uses an existing public benchmark and involves no crowdsourcing or direct interaction with human participants.

\item {\bf Institutional review board (IRB) approvals or equivalent for research with human subjects}
    \item[] Question: Does the paper describe potential risks incurred by study participants, whether such risks were disclosed to the subjects, and whether Institutional Review Board (IRB) approvals (or an equivalent approval/review based on the requirements of your country or institution) were obtained?
    \item[] Answer: \answerNA{}
    \item[] Justification: No new human-subject study, recruitment, intervention, or collection of identifiable participant data was conducted.

\item {\bf Declaration of LLM usage}
    \item[] Question: Does the paper describe the usage of LLMs if it is an important, original, or non-standard component of the core methods in this research? Note that if the LLM is used only for writing, editing, or formatting purposes and does \emph{not} impact the core methodology, scientific rigor, or originality of the research, declaration is not required.
    \item[] Answer: \answerNA{}
    \item[] Justification: No large language model is used as a component of the classifiers, training objective, adversarial attacks, or evaluation protocol.
\end{enumerate}
\end{document}